\documentclass[lettersize,journal]{IEEEtran}
\usepackage{amsmath,amsfonts}
\usepackage{algorithmic}
\usepackage{algorithm}
\usepackage{array}
\usepackage{textcomp}
\usepackage{stfloats}
\usepackage{url}
\usepackage{verbatim}
\usepackage{graphicx}
\usepackage{cite}
\usepackage{bibentry}
\usepackage{amssymb}
\usepackage{ragged2e} 
\usepackage{booktabs,makecell, multirow, tabularx}
\usepackage{bm}
\usepackage{pifont}

\usepackage[colorlinks, linkcolor=blue, anchorcolor=blue,
citecolor=green, urlcolor=black]{hyperref}
\usepackage{color}
\definecolor{gray}{RGB}{195,195,195}

\usepackage{caption}
\usepackage{graphicx}
\usepackage{float} 
\usepackage[caption=false,font=footnotesize,labelfont=rm,textfont=rm]{subfig}

\begin{document}
\title{PARFormer: Transformer-based Multi-Task Network for Pedestrian Attribute Recognition}

\author{Xinwen Fan, Yukang Zhang, Yang Lu,~\IEEEmembership{Member,~IEEE} and Hanzi Wang,~\IEEEmembership{Senior Member,~IEEE}
        % <-this % stops a space
\thanks{This work was supported in part by the National Natural Science Foundation of China under Grant U21A20514, and Grant 62002302; and in part by the Natural Science Foundation of Fujian Province under Grant 2020J01005 (Corresponding author: Hanzi Wang. E-mail: hanzi.wang@xmu.edu.cn).}       
\thanks{X. Fan and Y. Zhang  contributed equally to this work.}
\thanks{X. Fan, Y. Zhang, Y. Lu and H. Wang are with the  Fujian Key Laboratory of Sensing and Computing for Smart City, School of Informatics, Xiamen University, Xiamen 361005, China (E-mail: fanxinwen@stu.xmu.edu.cn, zhangyk@stu.xmu.edu.cn, luyang@xmu.edu.cn, hanzi.wang@xmu.edu.cn).}
}

\markboth{Journal of \LaTeX\ Class Files,~Vol.~14, No.~8, August~2021}%
{Shell \MakeLowercase{\textit{et al.}}: A Sample Article Using IEEEtran.cls for IEEE Journals}

\IEEEpubid{0000--0000/00\$00.00~\copyright~2021 IEEE}

\maketitle

\begin{abstract}
Pedestrian attribute recognition (PAR) has received increasing attention because of its wide application in video surveillance and pedestrian analysis. Extracting robust feature representation is one of the key challenges in this task. The existing methods mainly use the convolutional neural network (CNN) as the backbone network to extract features. However, these methods mainly focus on small discriminative regions while ignoring the global perspective. 
To overcome these limitations, we propose a pure transformer-based multi-task PAR network named PARFormer, which includes four modules. 
In the feature extraction module, we build a transformer-based strong baseline for feature extraction, which achieves competitive results on several PAR benchmarks compared with the existing CNN-based baseline methods. 
In the feature processing module, we propose an effective data augmentation strategy named batch random mask (BRM) block to reinforce the attentive feature learning of random patches. 
Furthermore, we propose a multi-attribute center loss (MACL) to enhance the inter-attribute discriminability in the feature representations. 
In the viewpoint perception module, we explore the impact of viewpoints on pedestrian attributes, and propose a multi-view contrastive loss (MCVL) that enables the network to exploit the viewpoint information. 
In the attribute recognition module, we alleviate the negative-positive imbalance problem to generate the attribute predictions. 
The above modules interact and jointly learn a highly discriminative feature space, and supervise the generation of the final features. 
Extensive experimental results show that the proposed PARFormer network performs well compared to the state-of-the-art methods on several public datasets, including PETA, RAP, and PA100K. Code will be released at \url{https://github.com/xwf199/PARFormer}.
\end{abstract}


%% fig 1
\begin{figure}[t]
	\centering
	\includegraphics[width=8.5cm]{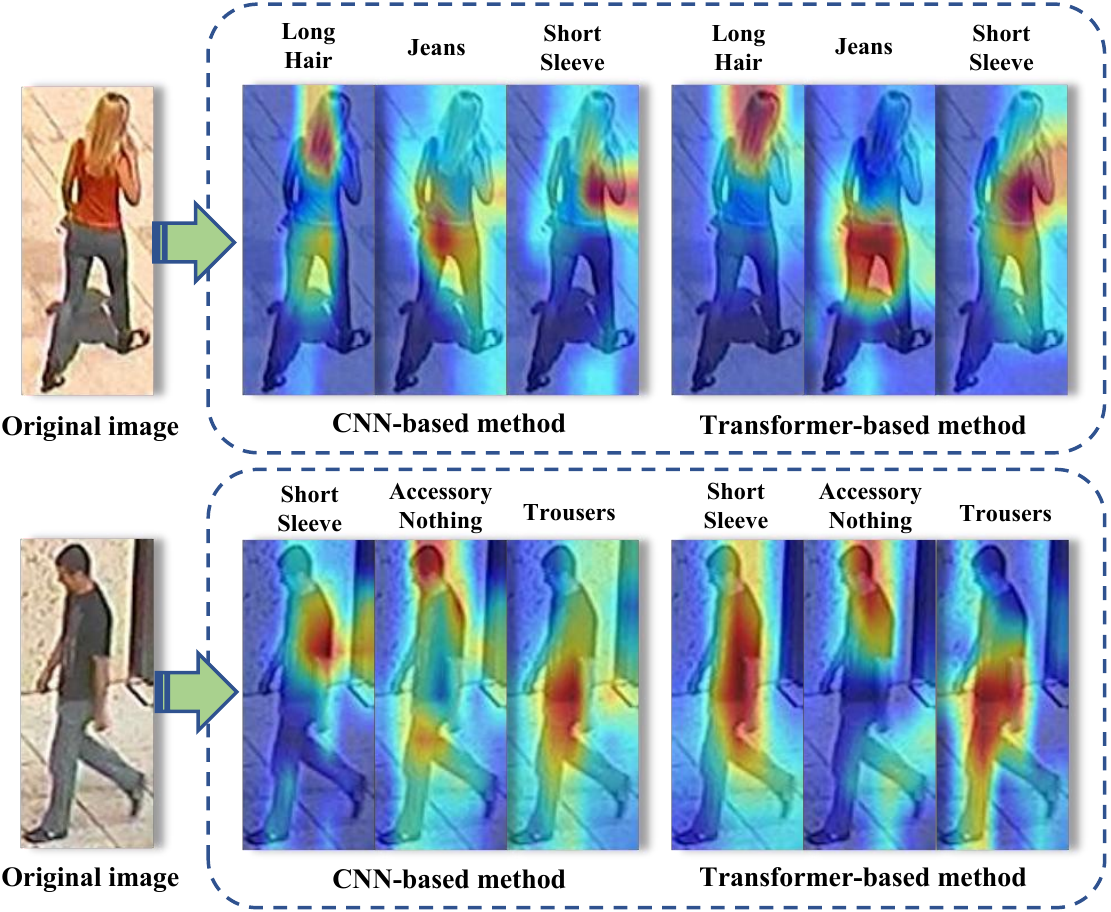}
	\caption{Comparison of the attribute heat maps generated by a CNN-based network (the middle column) with the proposed transformer-based baseline (the right column) for the PAR task. The transformer-based network, which combines the global perspective can capture more discriminative features. % (taking ResNet50 as an example)
	}
	\label{fig1}
\end{figure}

\section{Introduction}
\IEEEPARstart{P}{edestrian} attribute recognition (i.e., PAR) draws increasing attention due to its wide range of application scenarios. Given an image of a single person, the goal of PAR task is to recognize a set of semantic attributes (e.g., age, gender and clothing) to describe the characteristics of that pedestrian \cite{zhu2013pedestrian}. PAR has a great potential application in the field of intelligent video surveillance and security. It also be used as an auxiliary means for other tasks such as pedestrian detection \cite{lin2018multi, kim2021uncertainty, kumar2021diagonally}, pedestrian retrieval \cite{dong2019person, ke2022joint, jeong2021asmr} and pedestrian re-identification \cite{wang2020exploiting, zhang2021towards, zhang2023mrcn, chen2021explainable, zhang2023diverse}. PAR is a challenging task because the model is significantly affected by the viewpoint, light, and resolution \cite{wang2022pedestrian}.
%% pedestrian detection \cite{xiao2021deep, lin2018multi, kim2021uncertainty, kumar2021diagonally}, person retrieval \cite{dong2019person, fang2019bilinear, ke2022joint, jeong2021asmr} and pedestrian re-identification \cite{wang2020exploiting, zhang2021towards, li2021triple, chen2021explainable, zhao2019attribute}.

The existing PAR methods \cite{yaghoubi2020attention, guo2019visual} typically employ the convolutional neural network (CNN), such as ResNet \cite{he2016deep}, VGG \cite{simonyan2014very}, and InceptionNet \cite{szegedy2017inception, liu2018localization} as the backbone network. However, these CNN-based methods mainly focus on small discriminative regions and they may fail to exploit rich structural patterns at a global level. Recently, vision transformers (e.g., ViT \cite{dosovitskiy2020image}, DeiT \cite{touvron2021training} and Swin Transformer \cite{liu2021swin}) have been widely used for various computer vision tasks, which combine the learning methods of natural language processing and computer vision. 
%With the multi-head self-attention module, 
Using the multi-head self-attention module to focus on multiple areas at the same time, 
transformer-based models can capture long-range dependencies and retain more detailed information, which makes them suitable for extracting rich structural features. 
As shown in Fig. \textcolor{red}{1}, the proposed transformer-based network can obtain more discriminative features than the CNN-based network, which can extract more identification parts. 
\IEEEpubidadjcol

Although transformer can obtain features with a more global perspective, it still needs to be improved to accommodate changing environments in the PAR task. In the existing methods \cite{tang2019improving, han2019attribute}, the features extracted by the backbone network are deterministic. While in practical applications, the pedestrian images captured by cameras often contain challenges such as background clutter and occlusion, which may lead to the poor prediction of new samples by the trained model. The main reason is that external factors make certain local attributes hindered, so attentive feature learning for these areas is particularly important. 
% Therefore, we propose to mask the feature maps randomly, which enable the network to extract more robust features and adapt to changing factors during the training process.

Another problem with existing methods for feature processing is that the association between features of the same attribute sample is less explored. Existing loss functions \cite{zhang2018generalized, pang2019rethinking, lin2017focal, yang2021task} only consider the matching degree between the prediction results and the ground truth, while the center loss \cite{wen2016discriminative} provides a new strategy, which takes into account the similarity of samples in the same class, and brings them close together. However, each image only belongs to one class in the center loss, which cannot be applied to the PAR task. We draw inspiration from it to cluster samples of the same class, and aim to solve the feature association problem.

Moreover, exploration of the viewpoints in different images is also helpful for attribute recognition. There is a massive visual difference between images with different viewpoints, and the viewpoints have a great influence on attribute recognition task. Certain attributes are recognized better at specific viewpoints, such as glasses, collars, backpacks, etc. Perceiving the viewpoint information of images can enable the network to recognize attributes more accurately, and further improve the performance of the network. However, the viewpoint information is not full utilized by existing PAR methods.

Based on the superior performance of the transformer and overcoming the limitations of the aforementioned problems, we propose a pure transformer-based multi-task PAR network named PARFormer, which includes four modules, namely feature extraction, feature processing, viewpoint perception, and attribute recognition, respectively. 
We first build a transformer-based strong baseline for the feature extraction module. 
In order to extract robust feature representation, we propose a batch random mask (BRM) block in the feature processing module. This block enables the network to pay attention to different regions of the feature map and increase uncertainty in the training process, which makes it more robust to complex environments. Furthermore, we propose a multi-attribute center loss (MACL) to aggregate each attribute to their respective centers, and make the output features more discriminative. 
In the viewpoint perception module, we fully utilize the viewpoint information of samples, and propose a multi-view contrastive loss (MVCL), which enables the network to utilize the viewpoint information of pedestrian images to recognize attributes. 
In the attribute recognition module, we alleviate the negative-positive attribute imbalance problem to generate the attribute predictions more effectively. 
The above modules interact and contribute to the network, and jointly supervise the training process.

%The main contributions of this work can be summarized as the following:
The main contributions can be summarized as follows:
\begin{itemize}
	\item We propose a transformer-based multi-task network named PARFormer for the PAR task. We first build a strong baseline based on the transformer, and introduce a BRM block to increase the robustness of the network.
	\item We design a MACL that aggregates different attributes to their respective centers, and enables each attribute to be more effectively discriminated.
	\item We utilize the viewpoint information and propose a MVCL, by which the network can utilize the viewpoint information of pedestrian images to recognize attributes. 
	\item The experiments on three datasets (PETA, PA100K, and RAP) show that our network has excellent performance compared with several other state-of-the-art methods. In addition, we also conduct ablation experiments to verify the effectiveness of the proposed network.
\end{itemize}

%%%%%%%%%%%%%%%%%%%%%%%%%%%%%%%%%%%%%%%%%%%%%%%%%%%%%%%%%%%%
\section{Related Work}
\subsection{Pedestrian Attribute Recognition}
PAR has been widely researched in recent years. In the existing deep learning based PAR methods, some use auxiliary information, such as human key points \cite{2018Pose} and human parsing \cite{zhao2018grouping}, to assist training, while others do not use this information but only use attribute labels \cite{jia2021rethinking, li2019spatial, Sudowe2015person} in the training process.

\subsubsection{Methods based on auxiliary information}
These methods mainly use some auxiliary information to locate the specific region of an attribute, and then recognize the attribute based on this region. For example, Feng et al. \cite{2019Research} propose to use semantic segmentation to obtain the location information of the human body parts, and further study the attributes with known location information. Tang et al. \cite{tang2019improving} focus on the division of attribute-specific regions, and propose a attribute location module, which uses attribute information to adaptively find the most discriminative regions, and learns the regional features of each attribute at multiple levels. Zhao et al. \cite{zhao2018grouping} propose a grouping recurrent learning method, which detects the accurate body parts through a body region proposal network, and then extracts features from the detected regions. Li et al. \cite{2018Pose} propose a pose guided deep model, which extracts pose estimation from a pre-trained model, locates the body parts in the image, and uses multi-feature fusion based on location features for attribute recognition. 
Liu et al. \cite{liu2018localization} propose a localization guided network, which leverages class activation maps to obtain localization results, and uses the edge boxes \cite{zitnick2014edge} to obtain attribute specific local features. 
Although these methods achieve impressive performance, they require additional computation to obtain auxiliary information, and the performance of the model will be affected by the quality of auxiliary information.

%%% fig2
\begin{figure*}[t]
	\centering
	\includegraphics[width=17.5cm]{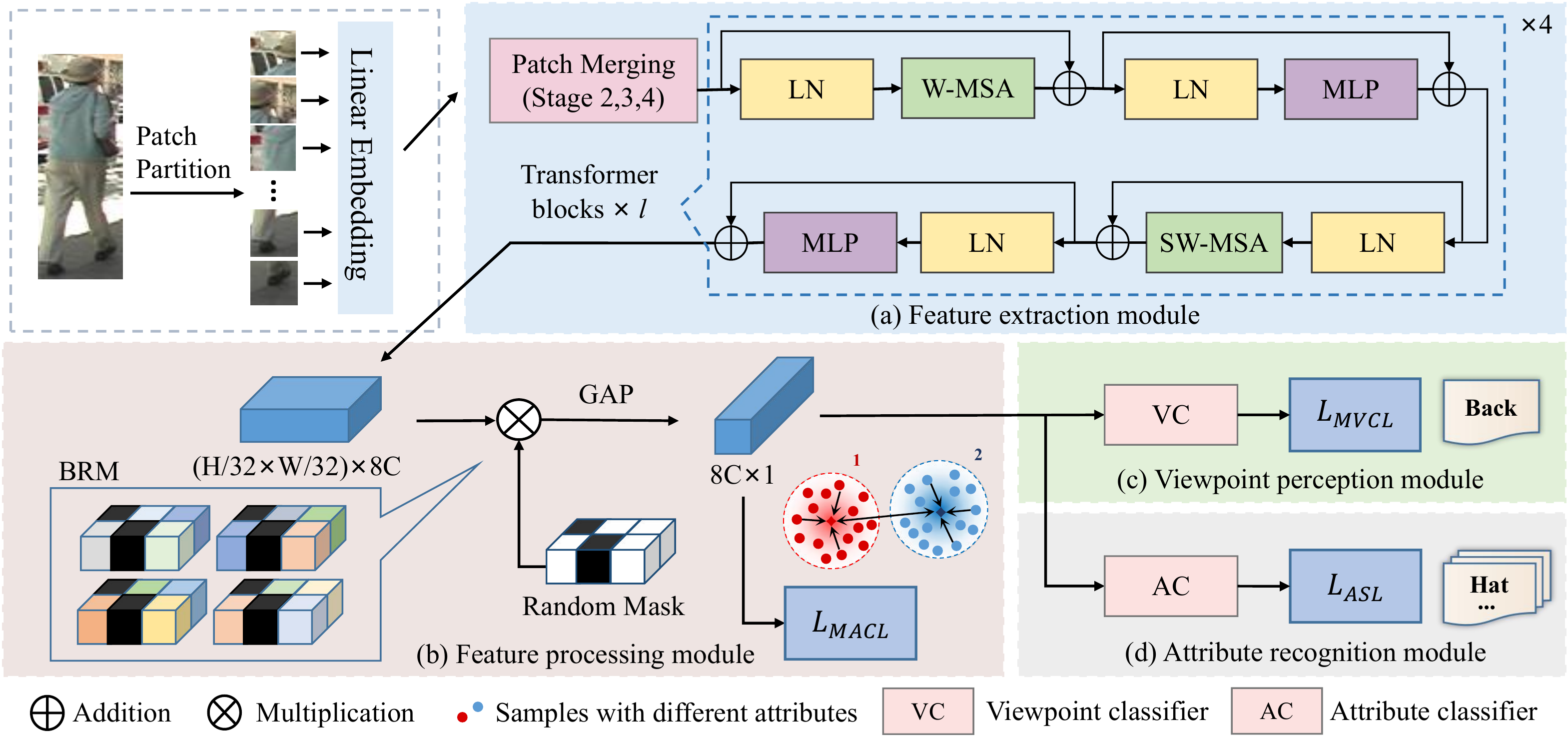}
	\caption{The overall network of the proposed PARFormer, which includes four modules: (a) Feature extraction module: The transformer-based backbone network extracts more discriminative features; (b) Feature processing module: The BRM block performs random mask on the feature map, and MACL aggregates features with the same attribute; (c) Viewpoint perception module: MVCL utilizes the viewpoint information to assist the attribute recognition; (d) Attribute recognition module: ASL alleviates the attribute imbalance problem and enables the network to perform better attribute prediction. }
	\label{fig1}
\end{figure*}

\subsubsection{Methods using only attribute labels}
These methods mainly focus on improving the performance of the network to obtain more discriminative features. Sarafianos et al. \cite{sarafianos2018deep} propose a multi-scale weakly supervised visual attention mechanism to extract attribute information, and a solution to deal with the category imbalance. Chen et al. \cite{chen2021explainable} propose an explainable network to generate the quantitative contribution of attributes and visualize the attention maps of the most discriminative attributes. Han et al. \cite{han2019attribute} explore the correlation of different attributes and propose a multi-branch architecture, and then collect the contextual information extracted from these branches to recognize the attributes. Zeng et al. \cite{zeng2020multi} propose a novel co-attentive sharing module to extract the channels and spatial regions, which enables the feature sharing effective in multi-task learning. 
Tan et al. \cite{tan2020relation} propose to adopt the graph convolutional network (GCN) \cite{welling2016semi} for PAR, which first generates attribute specific features, and then employs the GCN to learn the relationship between attributes. The method proposed by Nguyen et al. \cite{nguyen2021graph} takes pedestrian attribute labels and features as person signatures to form a graph, and then employs the GCN to learn the topological structure of pedestrian visual features. 
For the problem of incremental attributes, Xiang et al. \cite{xiang2019incremental} propose a meta-learning method to solve this problem, which can decompose information of multiple attributes and generalize new attributes. 
Although these methods achieve great success, such CNN-based methods only process one local neighborhood, and fail to exploit rich structural patterns at a global level.

\subsection{Transformer-Based Method}

\subsubsection{Vision transformer}
Transformer always plays a very important role in deep learning. It is a model proposed by Vaswani et al. \cite{2017Attentionis} for the natural language processing (NLP) task and it has a far-reaching impact. The vision transformer (ViT) proposed by Dosovitskiy et al. \cite{dosovitskiy2020image} makes a breakthrough by using the transformer in the field of computer vision. 
DeiT proposed by Touvron et al. \cite{touvron2021training} uses knowledge distillation to guide transformer to learn more effectively, which makes the network no longer need large-scale pre-training data. To solve the problem that ViT requires extensive computations, Liu et al. propose the swin transformer \cite{liu2021swin}, which introduces a shift window-based attention mechanism to reduce the computational cost.

\subsubsection{Application in the visual task}
In recent years, some researchers propose some transformer based methods in the visual task, which improve the performance and achieve better results. He et al. \cite{he2021transreid} propose to apply transformer to the re-identification of persons and objects. This method proves that transformer has great potential in ReID tasks. 
Lanchantin et al. \cite{lanchantin2021general} propose to apply transformer to multi-label image classification to exploit the dependencies among visual features and labels. Wang et al. \cite{wang2022pose} propose a transform based pose-guided feature disentangling method to solve the occlusion problem in person re-identification. The method proposed by Cheng et al. \cite{cheng2022simpl} encodes pedestrian images and attributes as visual and textual features, respectively. The proposed method performs cross-modal information interaction, and the transformer is served as a cross-modal fusion module. 
Xu et al. \cite{xu2022bridging} propose a two-step transformer based polishing network for video captioning. 
Dai et al. \cite{dai2021transformer} introduce a transformer-based structure to understand the distribution of instances for cross-view geo-localization.
Inspired by these, we propose to apply the transformer network to the PAR task.

%%%%%%%%%%%%%%%%%%%%%%%%%%%%%%%%%%%%%%%%%%%
%%%%%%%%%%%%%%%%%%%%%%%%%%%%%%%%%%%%%%%%%%%
%%%%%%%%%%%%%%%%%%%%%%%%%%%%%%%%%%%%%%%%%%
%%%%%%%%%%%%%%%%%%%%%%%%%%%%%%%%%%%%%%%%%%%%

\section{Proposed Network}
\subsection{Overall Network}
The proposed PARFormer is a multi-task network, as shown in Fig. \textcolor{red}{2}. The input of the proposed PARFormer is first splited into non-overlapping patches by the patch partition operation, and then the patches are fed into the multi-task network for feature learning. 

The proposed PARFormer includes four modules. In the feature extraction module, we build the backbone network based on the transformer, as shown in Fig. \textcolor{red}{2 (a)}. Our backbone consists of four stages and each stage has a different number of blocks. Before stages 2, 3, and 4, it performs the patch merging operation to build a hierarchical structure to obtain multi-scale features.

Then, we propose the BRM block and MACL in the feature processing module to learn a highly discriminative feature space. The global average pooling (GAP) layer is performed for the masked features after the BRM block to reduce the dimension. Afterward, the features are fed into MACL to cluster and enhance the distinction between different attributes. The red and blue dots in Fig. \textcolor{red}{2 (b)} represent the samples of a certain type of attribute, respectively, and MACL aggregates them to their respective centers. 

There are two classifiers within PARFormer, they are the viewpoint classifier (VC) and the attribute classifier (AC). The VC and AC belong to the viewpoint perception module and the attribute recognition module, respectively, as shown in Fig. \textcolor{red}{2 (c)} and Fig. \textcolor{red}{2 (d)}. The viewpoint information obtained from VC is used in MVCL to cluster the samples with  same viewpoint. The attribute recognition is supervised by asymmetric loss (ASL) to overcome the attribute imbalance.

\subsection{Feature Extraction}
For the feature extraction module, we build a PAR backbone network based on the transformer, and improve the backbone to obtain more discriminating features, as shown in Fig. \textcolor{red}{2 (a)}. Denote an input RGB image $x \in \mathbb{R}^{H\times W \times3} $, where $H$ and $W$ represent the height and width of the image, respectively, it is first splited into non-overlapping patches by the patch partition operation. We set the patch size to 4$\times$4 in the experiment. Then, through a linear embedding layer, the raw-valued feature is projected to an specific dimension (denoted as $C$).

The backbone network consists of four stages, and each stage contains several transformer blocks. In our experiment, these four stages contain 2, 2, 18, and 2 blocks, respectively. At the beginning of stages 2, 3, and 4, there will be a patch merging operation, which combines the adjacent small patches into a large patch, and plays a role of downsampling. 
As shown in Fig. \textcolor{red}{3}, patches in the same location merged together, and then the number of channels doubled through a linear layer. Therefore, after this operation, the length and width of the image will become 1/2 of the original ones, and the number of channels will be doubled.

In Fig. \textcolor{red}{2 (a)}, there are two continuous blocks shown in the dotted line. It has the same basic structure as the vision transformer block \cite{dosovitskiy2020image}, except for the window-based multi-head self-attention (W-MSA) and the shifted window-based multi-head self-attention (SW-MSA) \cite{liu2021swin}. 
W-MSA computes the self-attention within the non-overlapping windows, which achieves the linear computational complexity of the network with respect to the image size. 
To achieve the mutual communication between the windows, SW-MSA computes the self-attention of adjacent non-overlapping windows, and learns their relationship. 
The LayerNorm (LN) layer is applied before each MSA and MultiLayer Perceptron (MLP), and residual connections are applied after each MSA and MLP.

After the fourth stage, the original image becomes a feature map with the size of $(H/32 \times W/32) \times 8C$ is fed into the next processing stage.

%%% fig 3 patchmerge
\begin{figure}[t]
	\includegraphics[width=8.7cm]{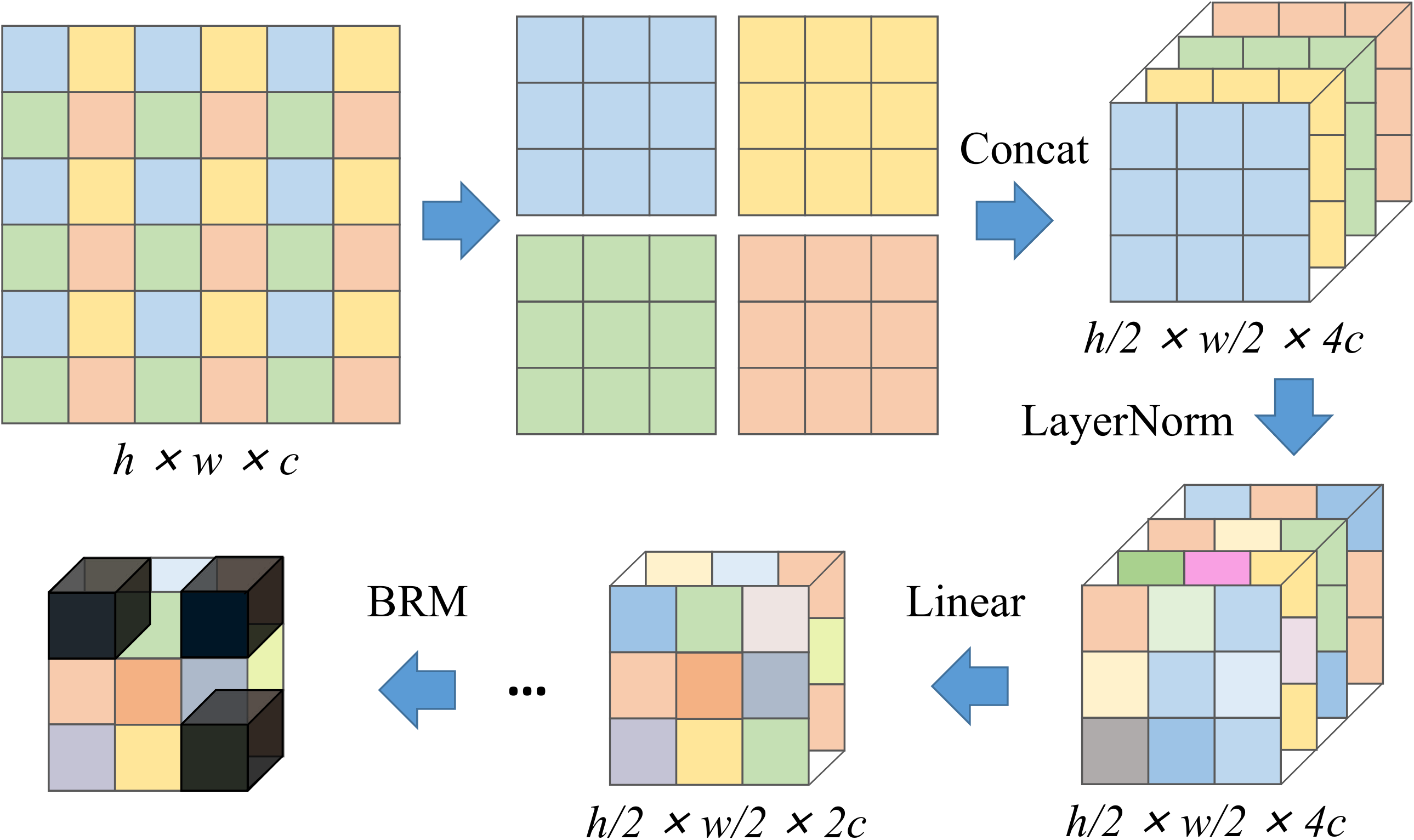}
	\caption{Patch merging and batch random mask (BRM) operation. The patch merging operation downsamples the image, and the adjacent patches of the image are moved to the same position in different channels. The pixels at the corresponding positions of these black parts are set to 0.}
	\label{fig1}
	\centering
\end{figure}

\subsection{Feature Processing}
In this module, we process the features extracted by the backbone network to make them more discriminative. We propose a novel BRM block to enhance the robustness of features, and propose an effective MACL to reduce differences between samples with the same attributes.

\subsubsection{Batch Random Mask Block}
In order to obtain more robust features and make the prediction of the network adapt to complex and changeable environments, we propose a batch random mask (BRM) block inspired by the Masked AutoEncoders (MAE) \cite{he2022masked}, in which the random patches of the image are masked to reconstruct these missing regions by the model. 
The MAE learns these unknown regions to reconstruct the masked patches, and achieves excellent results. 
In the proposed BRM block, we mask the feature map randomly, and reinforce the attentive feature learning of the other patches. This block randomly selects the patch locations on each batch, and masks the patches in the same locations of the feature maps in each minibatch. These masked patches are random and discontinuous. The box on the lower left in Fig. \textcolor{red}{2 (b)} indicates that different feature maps in the same batch are masked in the same region. This method can greatly enhance the robustness of the network.

The BRM block processes the feature maps output by the backbone network. The position of patches on these feature maps is greatly affected by the patch merging operation. As shown in Fig. \textcolor{red}{3}, the original feature size is $H \times W \times C$. First, elements are selected with position intervals of 2 in the row and column directions, and they are assembled into a new patch. Then, all the patches are concatenated into a complete tensor. 
At this time, $H$ and $W$ become 1/2 of the original size, and $C$ becomes four times of the original one. The linear layer is used to adjust the channel dimension $C$ to be twice of the original one. It can be seen that after the patch merging operation, the adjacent patches are disrupted, and the original adjacent patches will move to the same position in different dimensions on the feature map. In the next stage of the network, the information in each initial patch is retained in the corresponding position. 

The specific operation implemented in the BRM block is to randomly mask the 49 patches of the feature map in the last layer, i.e., to set the corresponding feature values to 0. 
These masked regions are variable, which are randomly assigned according to the mask ratio in each minibatch. The BRM block introduces extra perturbation in training, which enables the network to continuously learn and helps to generate more discriminative features.

In addition, the BRM block is directly added at the end of the backbone network. The reason is that the masked regions are evenly distributed in the feature map, and have no direct and obvious impact on the initial training process of the network. Therefore, this block is added to the end of the network and no additional branch is required.

\subsubsection{Multi-Attribute Center Loss}

% \usepackage{multirow}
% \usepackage[normalem]{ulem}

The center loss proposed by \cite{wen2016discriminative} is used to aggregate samples of the same class and reduce the intra-class differences in the facial expression recognition task. The formulation is as follows:
\begin{equation}
L_{c} = \frac{1}{2}  {\sum_{i=1}^{m}}  \left \| \bm{f}_{i}-\bm{c}_{y_{i}}  \right \| _{2}^{2},  
\end{equation}
where $m$ is the batch size, $\bm{f}_{i}$ represents the feature maps of image $i$ before the full connection layer, and $\bm{c}_{y_{i}}$ denotes the feature center of the $\bm{y}_{i}$ class.

However, each image only belongs to one class in the center loss, which cannot be used for the PAR task. Inspired by it, we propose a multi-attribute center loss (MACL) to reduce the differences within samples with the same attribute. We analogize the aggregation of similar samples to the aggregation of samples with the same attributes, i.e., aggregating the samples of each attribute to their respective centers. It increases the distance between attributes for better performance. 
MACL expects the features of each sample to be closer to the center of the specific attribute they possess:

\begin{equation}
\label{deqn_ex1a}
L_{MACL} = \frac{1}{2mn}  {\sum_{i=1}^{m}}  {\sum_{j=1}^{n}} \bm{y}_{ij}\cdot\left \| \bm{f}_{i}-\bm{c}_{y_{ij}}  \right \| _{2}^{2},  
\end{equation}
where $n$ is the number of attributes. $\bm{y}_{ij}$ represents the label of the $j$-th attribute in the image $i$, and it is used to limit that only the positive samples participate in the computation of MACL. The feature center $\bm{c}_{y_{ij}}$ is initialized randomly in each minibatch.

With the proposed MACL, the features extracted by the network are more discriminative.

%% Fig 4 viewlabel
\begin{figure}[t]
	\centering
	\includegraphics[width=8.6cm]{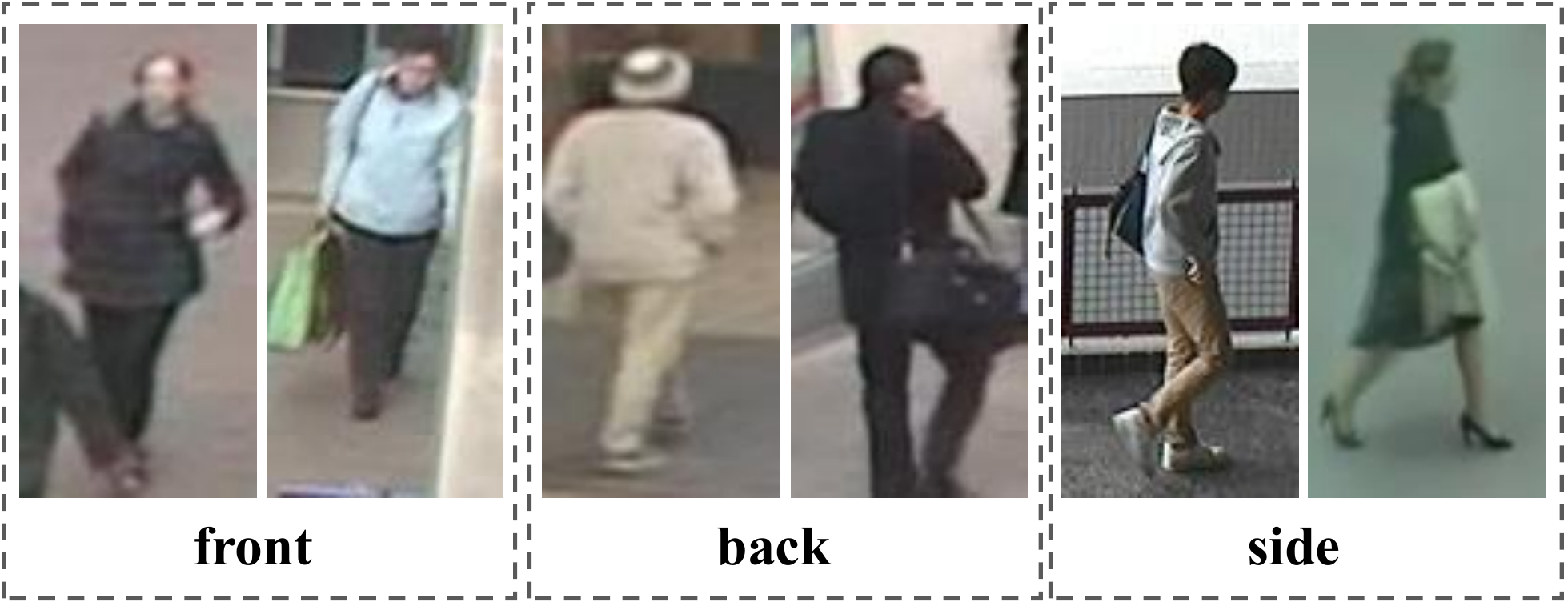}
	\caption{The viewpoint labels generated for the datasets include three types: front, back, and side. }
	\label{fig1}
\end{figure}

\subsection{Viewpoint Perception}
Different viewpoints will affect some specific attributes. Therefore, we design a viewpoint perception module to explore the viewpoint information and make full use of it. We generate viewpoint labels for the datasets and then propose a MVCL to cluster the samples with the same viewpoint.

\subsubsection{Viewpoint Label}
Li et al. \cite{li2016richly} explore the influence of different viewpoints on attribute recognition, and confirm that some attributes are easily recognized in certain viewpoints, which we call viewpoint-specific attributes. 
% For example, the recognition of the "backpack" attribute has the highest accuracy at the back, the recognition of the "glasses" attribute has the highest accuracy at the front, and the recognition of the "sleeve" attribute has the highest accuracy at the side. 
For example, the highest recognition accuracy for the "backpack" attribute is at the back viewpoint, for the "glasses" attribute is at the front viewpoint, and for the "sleeve" attribute is at the side viewpoint. 
To utilize the viewpoint information of images later, we add viewpoint labels to the datasets. 

For a pedestrian image in a PAR dataset, we define it as $I_{i}=(x_{i}, y_{i},v_{i})$, where $x_{i}$ is the RGB image and $y_{i}$ is the attribute label. We generate a viewpoint label $v_{i}$ for $I_{i}$, which is independent of $y_{i}$. 
Considering that the left and right viewpoints have similar effects on pedestrian attributes, we do not clearly distinguish between them, but only label them as the side viewpoint. Note that each image $x_{i}$ has only one viewpoint label $v_{i} \in (front; back; side) $, as shown in Fig. \textcolor{red}{4}. 

% We generate ground-truth viewpoint labels annotated by human for these two datasets, which consists of three labels: front, back, and side, as shown in Fig. \textcolor{red}{4}. 

% Due to the relatively ambiguous boundary of the viewpoint division, in our evaluation criteria, we specify that only images taken from the complete side are classified in the side class. Just as the first pedestrian image in Fig. \textcolor{red}{4} is taken from the left front, it is labeled as the front class.

\subsubsection{Multi-View Contrastive Loss}
After the feature processing module, we introduce a separate classifier to perceive the viewpoint of camera. We first calculate the cosine similarity of the pair representations, which reflects the correlation between samples. 
The cosine similarity of two samples $a$ and $b$ is calculated as follows:
\begin{equation}
sim(a,b)=\frac{a\cdot b}{\left \| a \right \| \times \left \| b \right \| }  =\frac{\sum_{j=1}^{n} (a_{j}\times b_{j} )}{\sqrt{\sum_{j=1}^{n}a_{j}^{2} }\times  \sqrt{\sum_{j=1}^{n}b_{j}^{2}}},
\end{equation}
where $a_{j}$ and $b_{j}$ denote the component of $a$ and $b$ on the $j$-th attribute, respectively. 
Then we propose a multi-view contrastive loss (MVCL) inspired by the contrastive learning \cite{2014Notes} for the viewpoint information:  
% , using the InfoNCE loss \cite{2014Notes}, which is widely used in self-monitoring learning:

\begin{equation}
L_{MVCL}(x_{i})= - \log \left \{ \frac{\exp \left[ sim(x_{i},x_{j}) /T \right]} { {\textstyle \sum_{k=1}^{m}  \exp \left[ sim(x_{i},x_{k}) /T \right]}} \right \} ,
\end{equation}
where $i \neq j$ and  $i \neq k$. $(x_{i}, x_{j})$ is a positive pair example, i.e., $v_{i}=v_{j}$. $T$ represents the temperature coefficient. 
% $s_{i,j}$ and $s_{i,k}$ correspond to the similarity of samples of the same and different viewpoints, respectively. 
MVCL compares the representations of pairwise samples to learn the viewpoint information implicitly, and then makes the samples of the same viewpoint closer after mapped to a lower dimensional space. In this way, the network can focus on learning some viewpoint-specific attributes and some attributes that are difficult to recognize from a certain viewpoint to improve recognition accuracy.

With MVCL, the network utilizes viewpoint information to assist the generation of features in the training process. Attribute information and viewpoint information jointly supervise the training process of the network, as shown in Fig. \textcolor{red}{5}. 
As a result, the network can more accurately recognize some view-specific attributes to achieve better performance. 
% Moreover, learning viewpoint information through MVCL introduce minimal extra computational cost. It is only executed during the training process, so it will not affect the efficiency in the inference stage. 

%% Fig 5:  attr_view
\begin{figure}[t]
	\centering
	\includegraphics[width=8.5cm]{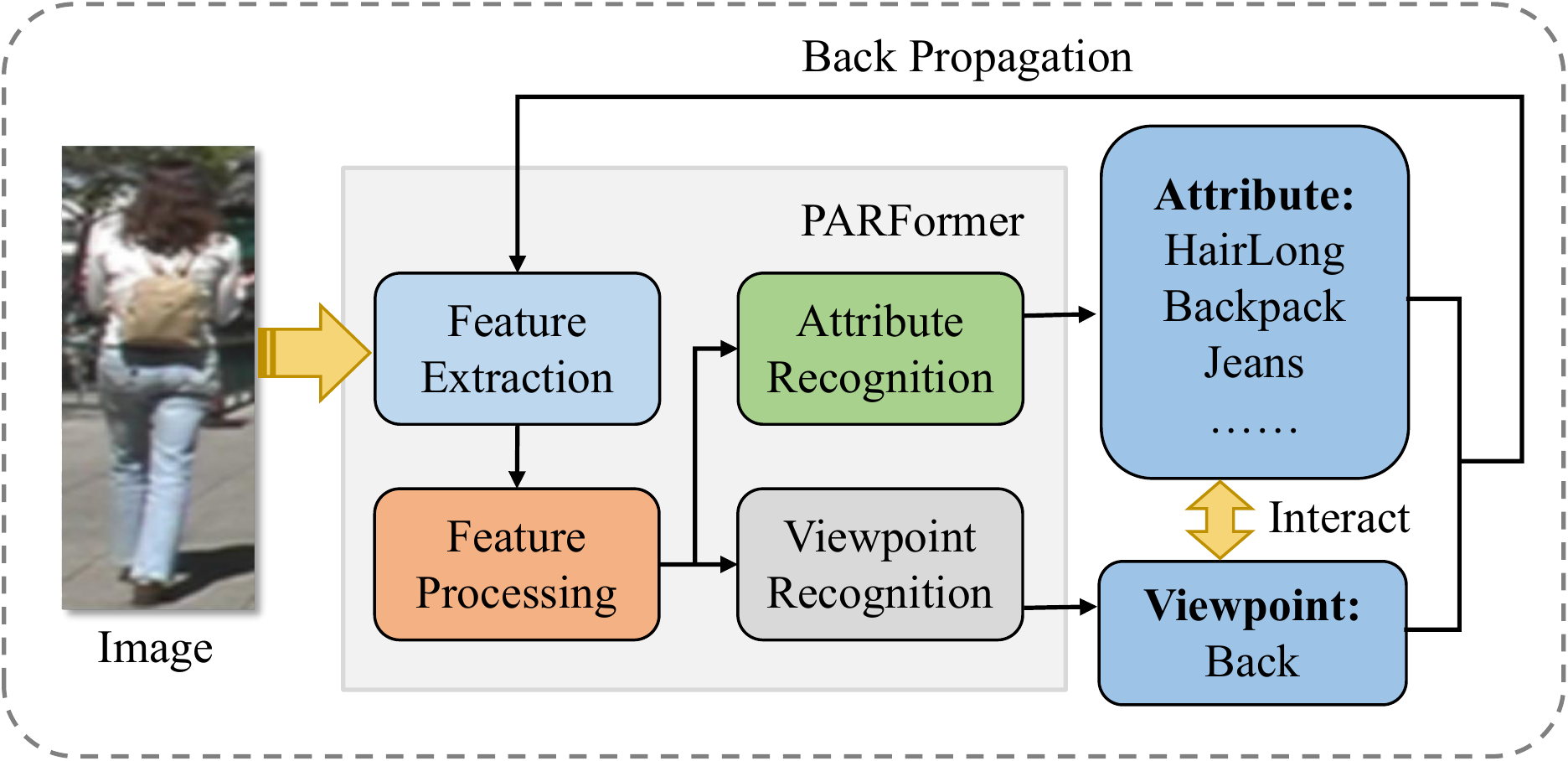}
	\caption{Attribute information and viewpoint information jointly supervise the feature extraction of PARFormer. }
	\label{fig1}
	\centering
\end{figure}

\subsection{Attribute Recognition}
Asymmetric loss (ASL) \cite{ridnik2021asymmetric} is an improved version of focal loss \cite{lin2017focal} to overcome the problem of imbalance between the positive and negative samples. We propose to apply it to the PAR task to alleviate the problem of attribute imbalance. The formulation is as follows: 
\begin{equation}
\begin{split}
L_{ASL} &= -{\textstyle \sum_{i=1}^{m}}  {\textstyle \sum_{j=1}^{n}} \bm{y}_{i,j}L_{+}+(1-\bm{y}_{i,j})L_{-},\\
\quad L_{+} &=(1-\bm{p}_{i,j})^{\gamma _{+}} \log(\bm{p}_{i,j}),\\
L_{-} &=(\bm{p}_{i,j})^{\gamma _{-}} \log(1-\bm{p}_{i,j}),
\end{split}
\end{equation}
where $\bm{y}_{ij}$ and $\bm{p}_{ij}$ represent the label and prediction probability of the $j$-th attribute in the image $i$. $\gamma _{+}$ and $\gamma _{-}$ correspond to the positive and negative focusing parameters, respectively.
ASL achieves a better balance between positive and negative
samples, thus improving the performance of the network.

\subsection{Multi-Loss Optimization}
Since the network parameters are shared, the loss functions of all modules are optimized jointly and concurrently to fully explore the pedestrian attribute information. 
MACL makes the output features more discriminative, while MVCL enables the network to perceive the viewpoint of the camera, and provides auxiliary information for the generation of features. 
Finally, the total loss of training the PARFormer network can be represented as: 
\begin{equation}
L = L_{ASL} + \lambda_{1} L_{MACL} + \lambda_{2} L_{MVCL},
\end{equation}
where the parameters $\lambda_1$ and $\lambda_2$ are used to control the specific gravity of the three loss functions.

%%%%%%%%%%%%%%%%%%%%%%%%%%%%%%%%%%%%%%%%%%%%
%%%%%%%%%%%%%%%%%%%%%%%%%%%%%%%%%%%%%%%%%%%%
%%%%%%%%%%%%%%%%%%%%%%%%%%%%%%%%%%%%%%%%%%%%

\section{Experiment}

\subsection{Datasets and Evaluation Metrics}
We evaluate the proposed methods on three widely used datasets, and adopt standard data split settings.

\textbf{PETA:} It contains a total of $19,000$ pedestrian images taken by real surveillance cameras \cite{deng2014pedestrian}. These images are randomly divided into $9,500$ training images, $1,900$ velidation images, and $7,600$ test images. Each pedestrian image has 61 binary attributes and 4 multi-category attributes. Because the distribution of some attributes is very uneven, the existing methods mainly focus on 35 attributes of the 61 attributes.

\textbf{PA100K:} It is the largest open-source pedestrian attribute dataset, with 26 pedestrian attributes annotated \cite{liu2017hydraplus}. 
It contains $100,000$ pedestrian images collected by the surveillance cameras, with $80,000$ images for training, $10,000$ images for validation, and $10,000$ images for testing.

\textbf{RAP:} It has two versions, and the RAP-v1 dataset \cite{li2016richly} is used in our experiment. This dataset contains $41,585$ pedestrian images collected from 26 indoor surveillance cameras, including 69 binary attributes, while the existing methods mainly focus on 51 attributes, and each of those is with a proportion greater than 1\%. The training set of this dataset contains $33,268$ images, and the rest are used for testing.

According to the existing methods, we adopt five metrics for evaluation: mean average precision (mA), accuracy (Accu), precision (Prec), recall (Recall), and F1 score (F1).

\iffalse
%% lamda
\begin{figure}[t]
	\centering
	\includegraphics[width=8.5cm]{lamda}
	\caption{Influence of $\lambda_1$ on the performance of the proposed method. }
	\label{fig1}
	\centering
\end{figure}
\fi

\subsection{Implementation Details}
To trade off the computation complexity and performance, we use both the swin-B and swin-L pre-trained on the imagenet-22k \cite{deng2009imagenet} for the backbone network. 
The corresponding network is abbreviated as PARFormer-B and PARFormer-L in the following, a difference between these two network is that the number of channels is 128 and 192, respectively.
In the four stages, the number of iterations for the transformer blocks is 2, 2, 18, and 2. 
Followed by \cite{2022stdp}, the size of the input image is resized to $3 \times 224 \times 224$ at the initial stage of the network training. 
The augmentations utilized include padding, random cropping, and random horizontal flipping. 
The batchsize is set to 32, and the training stops at 100 epochs. 
AdamW \cite{loshchilov2017decoupled} optimizer is adopted with a momentum of 0.9 and the weight decay of 0.05. The initial learning rate is set to $1 \times 10^{-4}$, and is decreased using the cosine decay schedule. The learning rate is halved in each cycle and reaches a minimum value of $5 \times 10^{-6}$ at the final epoch. 

For the viewpoint label, the RAP dataset provides viewpoint labels for each image, which include front, back, left, and right. The left and right viewpoints are merged into the side viewpoint. The PA100K dataset provides three viewpoint labels for each image: front, back, and side. 
For the PETA dataset, there are no viewpoint labels available. We generate the ground-truth viewpoint labels for this dataset, which consist of three labels: front, back, and side.

To investigate the effect of the hyperparameter, we give a rounded analysis in Section $ IV. D$. 
We follow the suggestion of \cite{ridnik2021asymmetric}, and we set $\gamma _{+}$ to 0 and $\gamma _{-}$ to 1 in the experiments on the three datasets. 
The other hyperparameter, mask ratio in the BRM block, temperature coefficient $T$ in Eq. (4), and the weight parameter $\lambda_{1}$ for MACL in Eq. (6) are set to (0.3, 0.1, 0.2) on the PETA dataset, (0.2, 0.2, 0.4) on the RAP dataset, and (0.2, 0.1, 0.5) on the PA100K dataset, respectively.  

\subsection{Comparative Results}
To show the effectiveness of the proposed network, we compare it with several state-of-the-art PAR methods: 
%These methods are quite different in practice, and have achieved superior results, which have high reference value. 
HPNet \cite{liu2017hydraplus}, VeSPA \cite{sarfraz2017deep}, JRL \cite{wang2017attribute}, 
PGDM \cite{2018Pose}, MsVAA \cite{sarafianos2018deep}, GRL \cite{zhao2018grouping}, 
VAC \cite{guo2019visual}, JLPSL-PAA \cite{tan2019attention}, RA \cite{zhao2019recurrent}, ALM \cite{tang2019improving},  
MT-CAS \cite{zeng2020multi}, DTM+AWK \cite{zhang2020deep}, JLAC \cite{tan2020relation}, 
Baseline\cite{jia2021rethinking}, SSC \cite{jia2021spatial}, DAFL \cite{jia2022learning}, and VTB\cite{cheng2022simpl}.
% SSR \cite{li2019spatial},  PD-Net \cite{liu2018localization}

%% PETA
\begin{table}
	\renewcommand\arraystretch{1.15}
	\centering
	\caption{Comparison with state-of-the-art methods on the PETA dataset. The \textcolor{red}{red} font and \textcolor{blue}{blue} font represent the highest and second highest scores, respectively. }
	\label{tab1}
	\begin{tabular}{c|ccccc}
		\hline
		\multirow{2}{*}{Method} & \multicolumn{5}{c}{PETA}                                                                                                                                                                                                                                                                                        \\   \cline{2-6}
		& \multicolumn{1}{c}{mA}             & \multicolumn{1}{c}{Accu}           & \multicolumn{1}{c}{Prec}           & \multicolumn{1}{c}{Recall}         & F1        \\ \hline 
		HPNet(ICCV17)         & \multicolumn{1}{c}{81.77}             & \multicolumn{1}{c}{76.13}             & \multicolumn{1}{c}{84.92}             & \multicolumn{1}{c}{83.24}             & 84.07         \\ 
		VeSPA(BMVC17)             & \multicolumn{1}{c}{83.45}             & \multicolumn{1}{c}{77.73}             & \multicolumn{1}{c}{86.18}             & \multicolumn{1}{c}{84.81}             & 85.49         \\
		JRL(ICCV17)          & \multicolumn{1}{c}{85.67}             & \multicolumn{1}{c}{--}             & \multicolumn{1}{c}{86.03}             & \multicolumn{1}{c}{85.34}             & 85.42         \\
		PGDM(ICME18)             & \multicolumn{1}{c}{82.97}             & \multicolumn{1}{c}{78.08}             & \multicolumn{1}{c}{86.86}             & \multicolumn{1}{c}{84.68}             & 85.76    \\ 
		MsVAA(ECCV18)          & \multicolumn{1}{c}{84.59}             & \multicolumn{1}{c}{78.56}             & \multicolumn{1}{c}{86.79}             & \multicolumn{1}{c}{86.12}             & 86.46    \\
		GRL(IJCAI18)          & 
		\multicolumn{1}{c}{86.70}          & \multicolumn{1}{c}{--}          & \multicolumn{1}{c}{84.34}          & \multicolumn{1}{c}{88.82}          & 86.51         \\ 
		% SSR(VCIP19)        & \multicolumn{1}{c}{84.28}          & \multicolumn{1}{c}{78.31}          & \multicolumn{1}{c}{86.52}          & \multicolumn{1}{c}{85.97}          & 86.24        \\ 
		JLPLS-PAA(TIP19)        & \multicolumn{1}{c}{84.88}          & \multicolumn{1}{c}{79.46}          & \multicolumn{1}{c}{87.42}          & \multicolumn{1}{c}{86.33}          & 86.87        \\ 
		RA(AAAI19)             & \multicolumn{1}{c}{86.11}          & \multicolumn{1}{c}{--}          & \multicolumn{1}{c}{84.69}          & \multicolumn{1}{c}{88.51}          & 86.56        \\ 
		ALM(ICCV19)             & \multicolumn{1}{c}{86.30}          & \multicolumn{1}{c}{79.52}          & \multicolumn{1}{c}{85.65}          & \multicolumn{1}{c}{88.09}          & 86.85          \\ 
		MT-CAS(ICME20)          & \multicolumn{1}{c}{83.17}          & \multicolumn{1}{c}{78.78}          & \multicolumn{1}{c}{87.49}          & \multicolumn{1}{c}{85.35}          & 86.41        \\
		% PD-Net(ICIP20)           & \multicolumn{1}{c}{84.85}          & \multicolumn{1}{c}{79.79}          & \multicolumn{1}{c}{87.99}          & \multicolumn{1}{c}{86.16}          & 87.07          \\
		DTM+AWK(20)             & \multicolumn{1}{c}{85.79}          & \multicolumn{1}{c}{78.63}          & \multicolumn{1}{c}{85.65}          & \multicolumn{1}{c}{87.17}          & 86.11          \\  
		JLAC(AAAI20)            & \multicolumn{1}{c}{86.96}          & \multicolumn{1}{c}{80.38}          & \multicolumn{1}{c}{87.81}    & \multicolumn{1}{c}{87.09}          & 87.45          \\ 
		Baseline (21)            & \multicolumn{1}{c}{85.17}          & \multicolumn{1}{c}{78.82}          & \multicolumn{1}{c}{86.77}          & \multicolumn{1}{c}{86.11}          & 86.44      \\ 
		SSC (ICCV21)            & \multicolumn{1}{c}{86.52}          & \multicolumn{1}{c}{78.95}          & \multicolumn{1}{c}{86.02}          & \multicolumn{1}{c}{87.12}          & 86.99    \\ 
		DAFL(AAAL22)            & \multicolumn{1}{c}{87.07}          & \multicolumn{1}{c}{78.88}          & \multicolumn{1}{c}{85.78}          & \multicolumn{1}{c}{87.03}          & 86.40               \\ \hline 
		PARFormer-B (Ours)                    & \multicolumn{1}{c}{\textcolor{blue}{\textbf{88.65}}}    & \multicolumn{1}{c}{\textcolor{blue}{\textbf{82.34}}}    & \multicolumn{1}{c}{\textcolor{blue}{\textbf{87.91}}}         & \multicolumn{1}{c}{\textcolor{blue}{\textbf{91.55}}}    & \textcolor{blue}{\textbf{88.66}}   \\ 
		PARFormer-L (Ours)                   & \multicolumn{1}{c}{\textcolor{red}{\textbf{89.32}}} & \multicolumn{1}{c}{\textcolor{red}{\textbf{82.86}}} & \multicolumn{1}{c}{\textcolor{red}{\textbf{88.06}}} & \multicolumn{1}{c}{\textcolor{red}{\textbf{91.98}}} & {\textcolor{red}{\textbf{89.06}}}\\ \hline
	\end{tabular} 
\end{table}

%% RAP
\begin{table}
	\renewcommand\arraystretch{1.15}
	\centering
	\caption{Comparison with state-of-the-art methods on the RAP dataset. The \textcolor{red}{red} font and \textcolor{blue}{blue} font represent the highest and second highest scores, respectively. }
	\label{tab1}
	\begin{tabular}{c|ccccc}
		\hline
		\multirow{2}{*}{Method}           & \multicolumn{5}{c}{RAP}                      \\   \cline{2-6}
		& \multicolumn{1}{c}{mA}     &\multicolumn{1}{c}{Accu}           & \multicolumn{1}{c}{Prec}           & \multicolumn{1}{c}{Recall}         & F1     \\ \hline 
		HPNet(ICCV17)         & \multicolumn{1}{c}{76.12}             & \multicolumn{1}{c}{65.39}             & \multicolumn{1}{c}{77.33}             & \multicolumn{1}{c}{78.79}             & 78.05         \\
		VeSPA(BMVC17)             & \multicolumn{1}{c}{77.70}             & \multicolumn{1}{c}{67.35}             & \multicolumn{1}{c}{\textcolor{blue}{\textbf{79.51}}}       & \multicolumn{1}{c}{79.67}             & 79.59     \\ 
		JRL(ICCV17)          & \multicolumn{1}{c}{77.81}             & \multicolumn{1}{c}{--}             & \multicolumn{1}{c}{78.11}             & \multicolumn{1}{c}{78.98}             & 78.58         \\
		PGDM(ICME18)              & \multicolumn{1}{c}{74.31}             & \multicolumn{1}{c}{64.57}             & \multicolumn{1}{c}{78.86}    & \multicolumn{1}{c}{75.90}             & 77.35   \\ 
		%LGNet(BMVC18)           & \multicolumn{1}{c}{78.68}             & \multicolumn{1}{c}{68.00}             & \multicolumn{1}{c}{80.36}    & \multicolumn{1}{c}{80.09}             & 77.55      \\
		% SSR(VCIP19)              & \multicolumn{1}{c}{79.92}             & \multicolumn{1}{c}{67.45}             & \multicolumn{1}{c}{80.46}    & \multicolumn{1}{c}{80.23}             & 80.34      \\
		JLPLS-PAA(TIP19)        & \multicolumn{1}{c}{81.25}          & \multicolumn{1}{c}{67.91}          & \multicolumn{1}{c}{78.56}          & \multicolumn{1}{c}{81.45}          & 79.98          \\ 
		RA(AAAI19)              & \multicolumn{1}{c}{81.16}          & \multicolumn{1}{c}{--}          & \multicolumn{1}{c}{79.45}          & \multicolumn{1}{c}{79.23}    & 79.34         \\
		ALM(ICCV19)             & \multicolumn{1}{c}{81.87}          & \multicolumn{1}{c}{68.17}          & \multicolumn{1}{c}{74.71}          & \multicolumn{1}{c}{86.48}    & 80.16          \\ 
		DTM+AWK(20)            & \multicolumn{1}{c}{82.04}          & \multicolumn{1}{c}{67.42}          & \multicolumn{1}{c}{75.87}          & \multicolumn{1}{c}{84.16}          & 79.80          \\  
		JLAC(AAAI20)           & \multicolumn{1}{c}{83.69}   & \multicolumn{1}{c}{69.15}          & \multicolumn{1}{c}{79.31}          & \multicolumn{1}{c}{82.40}          & 80.82          \\ 
		Baseline (21)            & \multicolumn{1}{c}{80.82}          & \multicolumn{1}{c}{67.56}          & \multicolumn{1}{c}{79.01}          & \multicolumn{1}{c}{80.37} & 79.68          \\ 
		SSC (ICCV21)            & \multicolumn{1}{c}{82.77}          & \multicolumn{1}{c}{68.37}          & \multicolumn{1}{c}{75.05}          & \multicolumn{1}{c}{87.49} & 80.43          \\ 
		DAFL(AAAI22)             & \multicolumn{1}{c}{83.72} & \multicolumn{1}{c}{68.18}          & \multicolumn{1}{c}{77.41}          & \multicolumn{1}{c}{83.39}          & 80.29          \\ 
		VTB(TCSVT22)             & \multicolumn{1}{c}{82.67}          & \multicolumn{1}{c}{69.44}          & \multicolumn{1}{c}{78.28}          & \multicolumn{1}{c}{84.39}          & 80.84    \\ \hline
		PARFormer-B (Ours)                  & \multicolumn{1}{c}{\textcolor{blue}{\textbf{83.84}}}          & \multicolumn{1}{c}{\textcolor{blue}{\textbf{69.70}}}    & \multicolumn{1}{c}{79.24}          & \multicolumn{1}{c}{\textcolor{blue}{\textbf{87.81}}}          & {\textcolor{blue}{\textbf{81.16}}}         \\ 
		PARFormer-L (Ours)                & \multicolumn{1}{c}{\textcolor{red}{\textbf{84.13}}}          & \multicolumn{1}{c}{\textcolor{red}{\textbf{69.94}}} & \multicolumn{1}{c}{\textcolor{red}{\textbf{79.63}}}    & \multicolumn{1}{c}{\textcolor{red}{\textbf{88.19}}}          & \textcolor{red}{\textbf{81.35}} \\ \hline
	\end{tabular} 
\end{table}

%% PA100K
\begin{table}
	\renewcommand\arraystretch{1.15}
	\centering
	\caption{Comparison with state-of-the-art methods on the PA100K dataset. The \textcolor{red}{red} font and \textcolor{blue}{blue} font represent the highest and second highest scores, respectively. }
	\label{tab1}
	\begin{tabular}{c|ccccc}
		\hline
		\multirow{2}{*}{Method}                                & \multicolumn{5}{c}{PA100K}     \\  \cline{2-6}
		& \multicolumn{1}{c}{mA}             & \multicolumn{1}{c}{Accu}           & \multicolumn{1}{c}{Prec}           & \multicolumn{1}{c}{Recall}         & F1       \\ \hline
		HPNet(ICCV17)         & \multicolumn{1}{c}{74.21}             & \multicolumn{1}{c}{72.19}             & \multicolumn{1}{c}{82.97}             & \multicolumn{1}{c}{82.09}             & 82.53         \\
		PGDM(ICME18)               & \multicolumn{1}{c}{74.95}          & \multicolumn{1}{c}{73.08}       & \multicolumn{1}{c}{84.36}    & \multicolumn{1}{c}{82.24}          & 83.29          \\ 
		% LGNet(BMVC18)           & \multicolumn{1}{c}{76.96}             & \multicolumn{1}{c}{75.55}             & \multicolumn{1}{c}{86.99}    & \multicolumn{1}{c}{83.17}             & 85.04      \\
		GRL(IJCAI18)          & \multicolumn{1}{c}{81.20}          & \multicolumn{1}{c}{--}       & \multicolumn{1}{c}{77.70}          & \multicolumn{1}{c}{80.90}          & 79.29       \\ 
		VAC(CVPR19)            & \multicolumn{1}{c}{79.16}          & \multicolumn{1}{c}{79.44}       & \multicolumn{1}{c}{\textcolor{red}{\textbf{88.97}}}    & \multicolumn{1}{c}{86.26}          & 87.59         \\
		JLPLS-PAA(TIP19)      & \multicolumn{1}{c}{81.61}          & \multicolumn{1}{c}{78.89}       & \multicolumn{1}{c}{86.83}          & \multicolumn{1}{c}{87.73}          & 87.27      \\ 
		ALM(ICCV19)             & \multicolumn{1}{c}{80.65}          & \multicolumn{1}{c}{77.08}       & \multicolumn{1}{c}{84.21}          & \multicolumn{1}{c}{88.84}          & 86.46 \\ 
		MT-CAS(ICME20)         & \multicolumn{1}{c}{77.20}          & \multicolumn{1}{c}{78.09}       & \multicolumn{1}{c}{\textcolor{blue}{\textbf{88.46}}}          & \multicolumn{1}{c}{84.86}          & 86.62   \\
		% PD-Net(ICIP20)       & \multicolumn{1}{c}{80.40}          & \multicolumn{1}{c}{78.80}       & \multicolumn{1}{c}{87.50}          & \multicolumn{1}{c}{86.91}          & 87.20 \\ 
		DTM+AWK(20)             & \multicolumn{1}{c}{81.63}          & \multicolumn{1}{c}{77.57}       & \multicolumn{1}{c}{84.27}          & \multicolumn{1}{c}{89.02}          & 86.58 \\  
		JLAC(AAAI20)            & \multicolumn{1}{c}{82.31}          & \multicolumn{1}{c}{79.47}       & \multicolumn{1}{c}{87.45}          & \multicolumn{1}{c}{87.77}          & 87.61  \\ 
		Baseline (21)           & \multicolumn{1}{c}{81.61}          & \multicolumn{1}{c}{79.45}       & \multicolumn{1}{c}{87.66}          & \multicolumn{1}{c}{87.59}          & 87.62   \\ 
		SSC (ICCV21)           & \multicolumn{1}{c}{81.87}          & \multicolumn{1}{c}{78.89}       & \multicolumn{1}{c}{85.98}          & \multicolumn{1}{c}{89.10}          & 86.87  \\ 
		DAFL(AAAI22)            & \multicolumn{1}{c}{83.54}          & \multicolumn{1}{c}{80.13}       & \multicolumn{1}{c}{87.01}          & \multicolumn{1}{c}{89.19}          & 88.09           \\ 
		VTB(TCSVT22)            & \multicolumn{1}{c}{83.72}    & \multicolumn{1}{c}{\textcolor{blue}{\textbf{80.89}}} & \multicolumn{1}{c}{87.88}          & \multicolumn{1}{c}{89.30}          & \textcolor{blue}{\textbf{88.21}}  \\ \hline
		PARFormer-B (Ours)                    & \multicolumn{1}{c}{\textcolor{blue}{\textbf{83.95}}}          & \multicolumn{1}{c}{80.26}       & \multicolumn{1}{c}{87.51}          & \multicolumn{1}{c}{\textcolor{blue}{\textbf{91.07}}}    & 87.69         \\ 
		PARFormer-L (Ours)                    & \multicolumn{1}{c}{\textcolor{red}{\textbf{84.46}}} & \multicolumn{1}{c}{\textcolor{red}{\textbf{81.13}}}       & \multicolumn{1}{c}{88.09}          & \multicolumn{1}{c}{\textcolor{red}{\textbf{91.67}}} & \textcolor{red}{\textbf{88.52}} \\ \hline
	\end{tabular} 
\end{table}

As shown in Table \textcolor{red}{I - III}, we compare the proposed PARFormer with the several existing methods. We use three public datasets and five evaluation metrics for comparative evaluation. 
The experimental results show that the PARFormer has excellent performance compared with the current optimal methods, and achieves the best mA, Accu, Recall, and F1 scores across all datasets. 
Additionally, the PARFormer-L network demonstrates better performance but greater computational cost compared with the PARFormer-B network. 

\textbf{PETA:} As shown in Table \textcolor{red}{I}, the proposed PARFormer-L outperforms the competitors on all five metrics with 89.32\% in mA, and the PARFormer-B achieves the second highest scores in all five metrics. The proposed PARFormer-L outperforms the previous best DAFL by 2.25\%, 3.98\%, 2.28\%, 4.95\%, and 2.66\% in the five evaluation metrics, respectively.

\textbf{RAP:} As shown in Table \textcolor{red}{II}, the proposed PARFormer-L outperforms the competitors on all five metrics. The PARFormer-B achieves the second highest mA, Accu, Recall and F1 values, and achieves a competitive Prec value which is better than VTB and DAFL. Compared with the previous best DAFL, the PARFormer-L improves 0.41\%, 1.76\%, 2.22\%, 4.80\%, and 1.06\% in the five evaluation metrics, respectively.

\textbf{PA100K:} As shown in Table \textcolor{red}{III}, the proposed PARFormer-L outperforms the competitors on the mA, Accu, Recall, and F1 score, and achieves a competitive Prec value second only to MT-CAS. The PARFormer-B achieves the second-best mA and Recall values, and achieves the third-best Accu value. The proposed PARFormer-L respectively improves 0.74\%, 0.24\%, 0.21\%, 2.37\%, and 0.31\% in the five evaluation metrics compared with the previous best VTB.

From the experimental performance on the three datasets, we can observe that PARFormer achieves excellent results. The exceptional performance of the PARFormer can be attributed to the transformer architecture, by which the network learn the global context and capture the long range dependencies in different regions. 
In addition, the BRM block strengthens the robustness of the network by reinforce the attentive feature learning, MACL clusters samples with the same attribute to their respective centers, and MVCL leverages viewpoint information to assist the training process. PARFormer integrates the BRM block, MACL and MVCL into a unified network that further enhance the network performance and make the predictions more competitive.

%% table 2: ablation
\begin{table*}[]
	\renewcommand\arraystretch{1.15}
	\centering
	\caption{Ablation study on the PETA, PA100K, and RAP datasets. Performance improvements validate the influence of each component of the proposed method. ASL: asymmetric loss; MACL: multi-attribute center loss; BRM: batch random mask; MVCL: multi-view contrastive loss. The symbols \ding{51} and \ding{55} indicate that the corresponding component is included or excluded. The \textbf{bold} numbers represent the best results under each case. }
	\label{tab1} %% 17.6  18.3
	\resizebox{\textwidth}{19mm}{
		\begin{tabular}{ c|c c c c|llll|llll|llll }
			\hline
			\multirow{2}{*}{Method} & \multirow{2}{*}{ASL}      & \multirow{2}{*}{MACL}       & \multirow{2}{*}{BRM}      & \multirow{2}{*}{MVCL}                            & \multicolumn{4}{c|}{PETA}                                                                                                                    & \multicolumn{4}{c|}{PA100K}                                                                                                               & \multicolumn{4}{c}{RAP}                                                                                                                  \\ \cline{6-17} 
			&                           &                           &                           &                                                & \multicolumn{1}{c}{mA}             & \multicolumn{1}{c}{Accu}           & \multicolumn{1}{c}{Recall}         & \multicolumn{1}{c|}{F1}    & \multicolumn{1}{c}{mA}             & \multicolumn{1}{c}{Accu}           & \multicolumn{1}{c}{Recall}         & \multicolumn{1}{c|}{F1} & \multicolumn{1}{c}{mA}             & \multicolumn{1}{c}{Accu}           & \multicolumn{1}{c}{Recall}         & \multicolumn{1}{c}{F1} \\ \hline 
			1                      & \textcolor{gray}{\ding{55}}         &  \textcolor{gray}{\ding{55}}      &  \textcolor{gray}{\ding{55}}    &   \textcolor{gray}{\ding{55}}                      & \multicolumn{1}{l}{86.85}          & \multicolumn{1}{l}{81.36}          & \multicolumn{1}{l}{87.64}          & 87.83                      & \multicolumn{1}{l}{81.89}          & \multicolumn{1}{l}{79.07}          & \multicolumn{1}{l}{87.17}          & 86.73                   & \multicolumn{1}{l}{80.21}          & \multicolumn{1}{l}{68.71}          & \multicolumn{1}{l}{82.23}          & 80.19                   \\ 
			2                      & \ding{51} & \textcolor{gray}{\ding{55}}    &   \textcolor{gray}{\ding{55}}     &    \textcolor{gray}{\ding{55}}                 & \multicolumn{1}{l}{87.52}          & \multicolumn{1}{l}{81.69}          & \multicolumn{1}{l}{90.43}          & 87.97                      & \multicolumn{1}{l}{82.57}          & \multicolumn{1}{l}{79.29}          & \multicolumn{1}{l}{89.06}          & 87.25                   & \multicolumn{1}{l}{81.86}          & \multicolumn{1}{l}{68.97}          & \multicolumn{1}{l}{86.82}          & 80.42                   \\
			3    & \textcolor{gray}{\ding{55}} & \ding{51}     &  \textcolor{gray}{\ding{55}} &  \textcolor{gray}{\ding{55}}                     & \multicolumn{1}{l}{87.56}          & \multicolumn{1}{l}{81.73}          & \multicolumn{1}{l}{89.36}          & 88.04                      & \multicolumn{1}{l}{82.38}          & \multicolumn{1}{l}{79.36}          & \multicolumn{1}{l}{88.30}          & 87.08                   & \multicolumn{1}{l}{81.64}          & \multicolumn{1}{l}{69.02}          & \multicolumn{1}{l}{84.23}          & 80.31                   \\ 
			4     & \textcolor{gray}{\ding{55}} & \textcolor{gray}{\ding{55}} &    \ding{51}            &  \textcolor{gray}{\ding{55}}                                   & \multicolumn{1}{l}{87.43}          & \multicolumn{1}{l}{81.87}          & \multicolumn{1}{l}{89.15}          & 87.95                      & \multicolumn{1}{l}{82.66}          & \multicolumn{1}{l}{79.62}          & \multicolumn{1}{l}{88.70}          & 87.13                   & \multicolumn{1}{l}{81.98}          & \multicolumn{1}{l}{69.25}          & \multicolumn{1}{l}{85.43}          & 80.55                   \\ 
			5 &\textcolor{gray}{\ding{55}}
			&\textcolor{gray}{\ding{55}} &\textcolor{gray}{\ding{55}} &\ding{51}                          & \multicolumn{1}{l}{87.77}          & \multicolumn{1}{l}{81.83}          & \multicolumn{1}{l}{89.96}          & 88.15                      & \multicolumn{1}{l}{82.70}          & \multicolumn{1}{l}{79.79}          & \multicolumn{1}{l}{88.58}          & 87.21                   & \multicolumn{1}{l}{82.04}          & \multicolumn{1}{l}{69.34}          & \multicolumn{1}{l}{85.21}          & 80.43                   \\ 
			6                      & \ding{51} & \ding{51} & \textcolor{gray}{\ding{55}} &  \textcolor{gray}{\ding{55}}     & \multicolumn{1}{l}{88.35}          & \multicolumn{1}{l}{81.85}          & \multicolumn{1}{l}{90.72}          & 88.12                      & \multicolumn{1}{l}{83.28}          & \multicolumn{1}{l}{79.54}          & \multicolumn{1}{l}{90.56}          & 87.36                   & \multicolumn{1}{l}{82.84}          & \multicolumn{1}{l}{69.10}          & \multicolumn{1}{l}{87.17}          & 80.75                   \\ 
			7                      & \ding{51} & \ding{51} & \ding{51} & \textcolor{gray}{\ding{55}}                                             & \multicolumn{1}{c}{88.43}          & \multicolumn{1}{c}{82.00}          & \multicolumn{1}{c}{91.33}          & \multicolumn{1}{c|}{88.34} & \multicolumn{1}{l}{83.69}          & \multicolumn{1}{l}{79.89}          & \multicolumn{1}{l}{90.80}          & 87.48                   & \multicolumn{1}{l}{83.19}          & \multicolumn{1}{l}{69.32}          & \multicolumn{1}{l}{87.53}          & 80.88                   \\  
			8                      & \ding{51} & \ding{51} & \ding{51} & \multicolumn{1}{c|}{\ding{51}} & \multicolumn{1}{l}{\textbf{88.65}} & \multicolumn{1}{l}{\textbf{82.34}} & \multicolumn{1}{l}{\textbf{91.55}} & \textbf{88.66}             & \multicolumn{1}{l}{\textbf{83.95}} & \multicolumn{1}{l}{\textbf{80.26}} & \multicolumn{1}{l}{\textbf{91.07}} & \textbf{87.69}          & \multicolumn{1}{l}{\textbf{83.84}} & \multicolumn{1}{l}{\textbf{69.70}} & \multicolumn{1}{l}{\textbf{87.81}} & \textbf{81.16}          \\ \hline
	\end{tabular}}
\end{table*}

\begin{figure*}[t]
	\centering
	\includegraphics[width=17.cm]{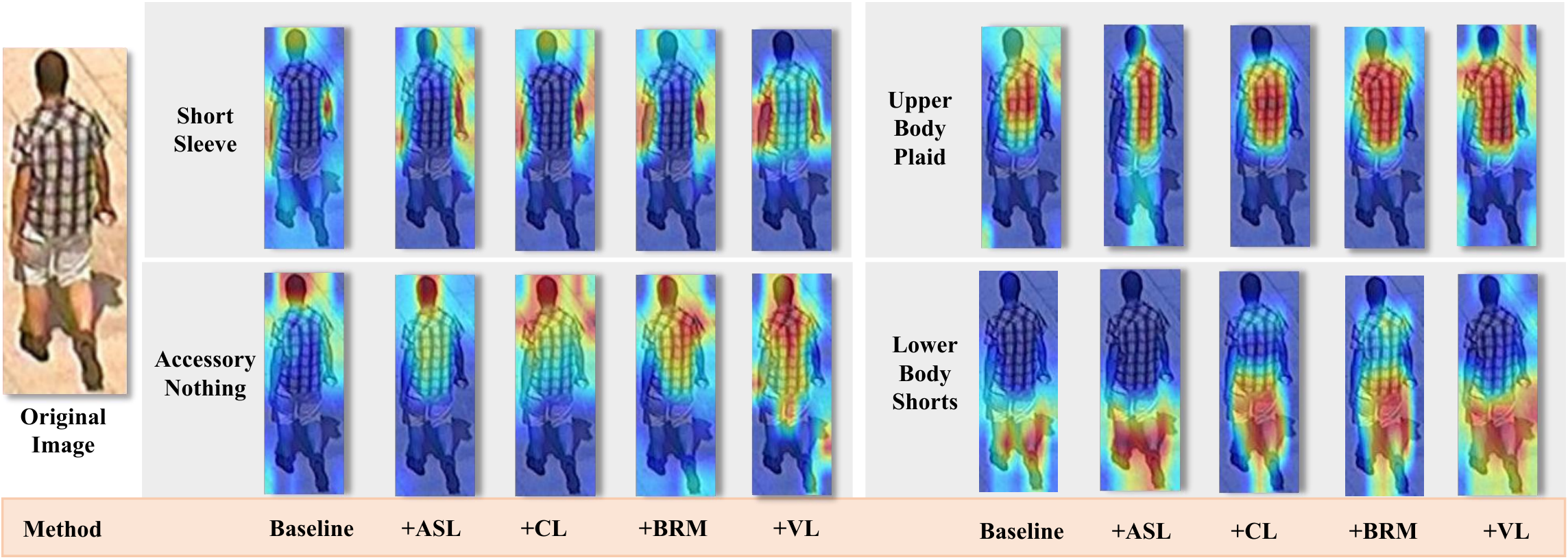}
	\caption{Grad-CAM \cite{selvaraju2017grad} visualization of attention heat maps. The methods from left to right indicate that the module or loss function is successively applied to show their cumulative effect. }
	\label{fig1}
	\centering
\end{figure*}

%% badkbone
\begin{table}[]
	\centering
	\renewcommand\arraystretch{1.15}
	\caption{Comparison of the CNN-based and transformer-based methods on the PA100K dataset. The \textcolor{red}{red} font and \textcolor{blue}{blue} font represent the first and second highest scores, respectively.}
	\label{tab1}
	\begin{tabular}{c|ccccc}
		\hline
		Method       & mA             & Accu           & Prec           & Recall         & F1             \\ \hline 
		TResNet-M \cite{ridnik2021tresnet}    & 74.25          & 68.13          & 79.77          & 80.05          & 80.16          \\ 
		BN-Inception \cite{ioffe2015batch} & 79.12          & 77.29          & 86.13          & 86.12          & 86.21          \\ 
		ResNet50 \cite{he2016deep}     & 79.41          & 78.05          & 86.84          & 86.75          & 86.59          \\ 
		ResNet101 \cite{he2016deep}    & 79.72          & 78.28          & \textcolor{red}{\textbf{87.05}} & 86.81          & \textcolor{blue}{\textbf{86.64}}          \\ \hline
		ViT-B \cite{dosovitskiy2020image}        & \textcolor{blue}{\textbf{81.11}}          & \textcolor{blue}{\textbf{78.35}}          & 86.62          & \textcolor{blue}{\textbf{86.89}}          & 86.52          \\
		Swin-B \cite{liu2021swin}      & \textcolor{red}{\textbf{81.89}} & \textcolor{red}{\textbf{79.07}} & \textcolor{blue}{\textbf{86.87}}          & \textcolor{red}{\textbf{87.17}} & \textcolor{red}{\textbf{86.73}} \\ \hline
	\end{tabular}
\end{table}

\iffalse
%% ratio
\begin{figure}[t]
	\centering
	\includegraphics[width=8.5 cm]{ratio}
	\caption{Influence of the mask ratio in the BRM block.}
	\label{fig1}
	\centering
\end{figure}
%% T
\begin{figure}[t]
	\centering
	\includegraphics[width=8.5 cm]{temp}
	\caption{Influence of the temperature coefficient $T$ in the MVCL.}
	\label{fig1}
	\centering
\end{figure}
\fi

\subsection{Model Evaluation}
\subsubsection{Ablation Study}
To better demonstrate the effectiveness and advantage of each employed component on the performance of the PARFormer network, we conduct ablation experiments on the three datasets. 

Table \textcolor{red}{IV} shows the results of ablation experiments on the PETA, PA100K, and RAP datasets. We take mA, Accu, recall, and F1 score to compare the results, given that Prec and Recall are usually share a negative correlation with each other. 
Method 1 only utilizes the backbone network with the binary cross entropy (BCE) loss, and we take it as the baseline for the following comparisons. Method 2 replaces the BCE loss with ASL, which achieves the improvement of performance by the effective balance of samples, 
and it brings 0.67\%, 0.68\%, 1.65\% improvements in mA on the three datasets, respectively.
Method 3 incorporates MACL into method 1, it improves the discrimination of features by clustering samples with the same attributes, and it brings 0.71\%, 0.49\%, 1.43\% higher values in mA on the three datasets, respectively. 
Method 4 incorporates the BRM block into method 1 to enhance feature learning and improve the robustness of the network, and it improves 0.58\%, 0.77\%, 1.77\% in mA on the three datasets, respectively. 
Method 5 integrates MVCL into method 1 to utilize the viewpoint information to assist the attribute recognition, and it improves the performance by 0.92\%, 0.81\%, 1.83\% in mA on the three datasets. 

Methods 6 to 8 combine these elements to investigate their cumulative effect, which further validates the effectiveness and superiority of the proposed methods. The results show that method 8 achieves the best results across all three datasets, in which the mA value on the three datasets are improved 1.80\%, 2.06\%, and 3.63\%, respectively. The experimental results show the advantages of applying ASL to the PAR task, and demonstrate the effectiveness of the proposed BRM block, MACL, MVCL, and the overall network.

The attention heat maps visualized in Fig. \textcolor{red}{6} qualitatively show the cumulative effect of the proposed modules and loss functions. The results show that the proposed network is able to focus on more discriminative image regions and make more accurate predictions when compared to the baseline.

\subsubsection{Analysis of the Backbone Network}
Table \textcolor{red}{V} shows the comparison of the CNN-based and transformer-based methods on the PA100K dataset. 
We perform experiments in the same environment for a fair comparison and set the batch size to 32. 
We can observe that the transformer-based methods such as vision transformer (ViT-B) and swin transformer (Swin-B) have higher performance than the CNN-based methods on the PAR task. 
Swin-B achieves higher performance than ViT-B with the aid of the unique shift-windows attention mechanism and hierarchical structure, which allows the network to capture more discriminative features.

\subsubsection{Analysis of the Proportion of Loss Function}
We use $\lambda_1$ and $\lambda_2$ in Equation (6) to control the proportion of the two loss functions. The weight $\lambda_2$ of MVCL is uniformly set to 1.0. 
We conduct experiments to investigate the effect of different values of the weight $\lambda_1$ of MACL. 
Fig. \textcolor{red}{7 (a)}, \textcolor{red}{(b)}, and \textcolor{red}{(c)} show the impact of $\lambda_1$ on the performance of the proposed PARFormer network. 
The highest performance is achieved when $\lambda_1$ are set to 0.2, 0.4, and 0.5 on the PETA, RAP and PA100K datasets, respectively. Among them, the $\lambda_1$ value on the PETA dataset is lower than that on the other two datasets. 
The reason is that the PETA dataset has the smallest scale and fewer attribute types, resulting in a relatively low demand for MACL.

%%%  Fig.7  参数实验
\begin{figure*}
	\begin{minipage}[t]{.3\linewidth}
		\centering
		\subfloat[][$\lambda_1$ on the PETA dataset]{\label{Genelecs:Genelec 8030 AP}\includegraphics[width=5cm]{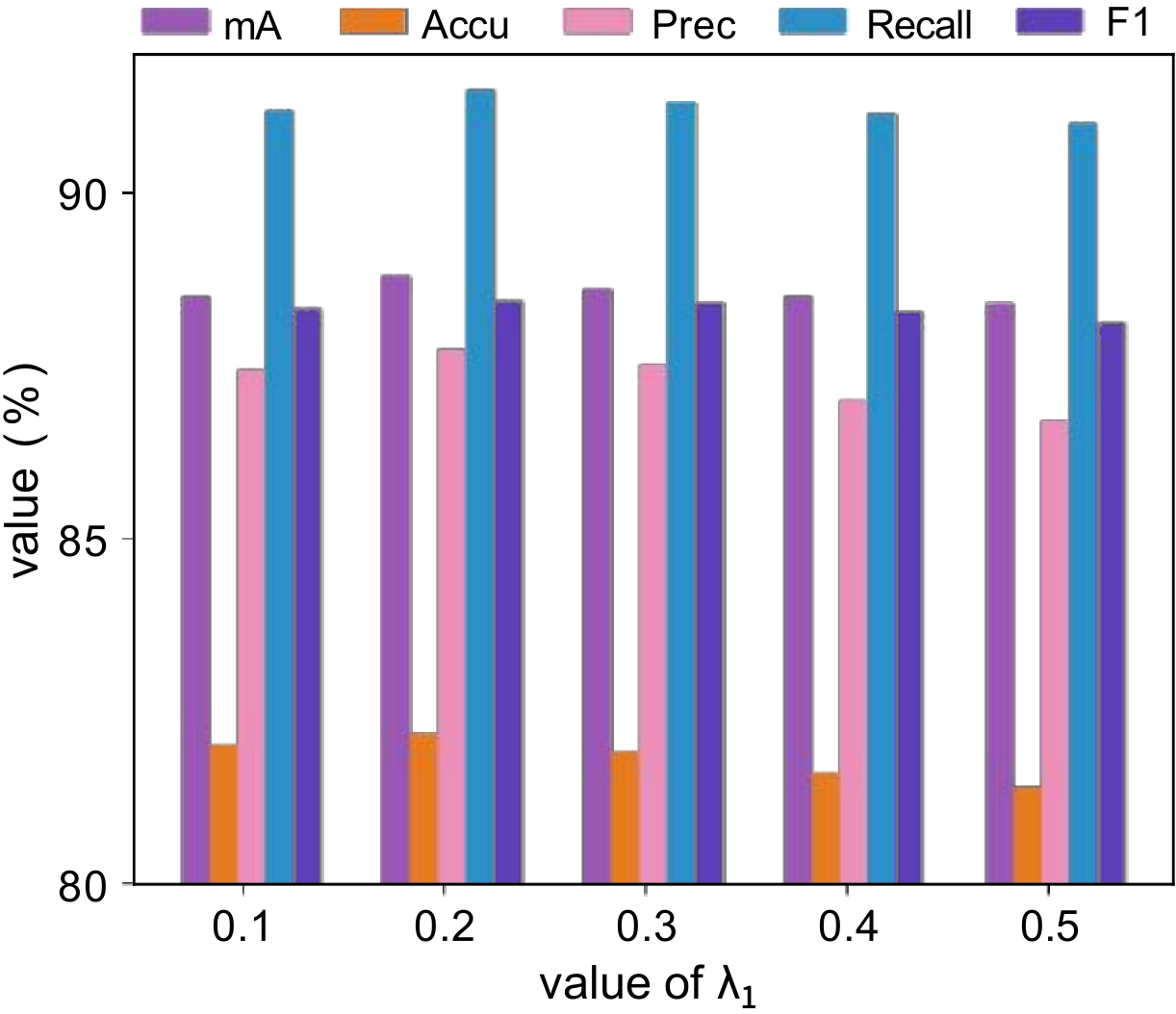}}
	\end{minipage} 
	\hspace{.12in}
	\begin{minipage}[t]{.3\linewidth}
		\centering
		\subfloat[][$\lambda_1$ on the RAP dataset]{\label{Genelecs:Genelec 8030 AP}\includegraphics[width=5cm]{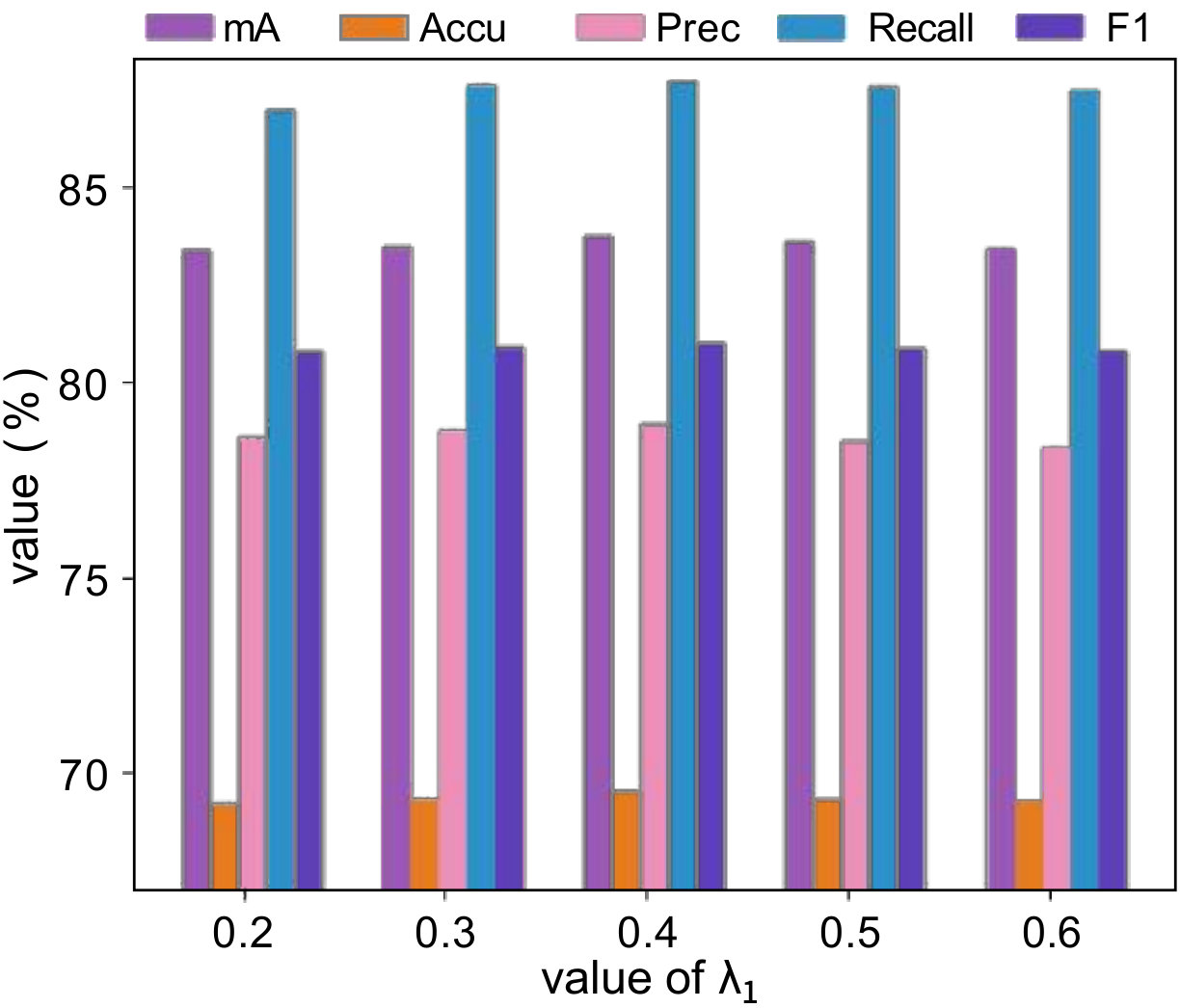}}
	\end{minipage} 
	\hspace{.12in}
	\begin{minipage}[t]{.3\linewidth}
		\centering
		\subfloat[][$\lambda_1$ on the PA100K dataset]{\label{Genelecs:Genelec 8030 AP}\includegraphics[width=5cm]{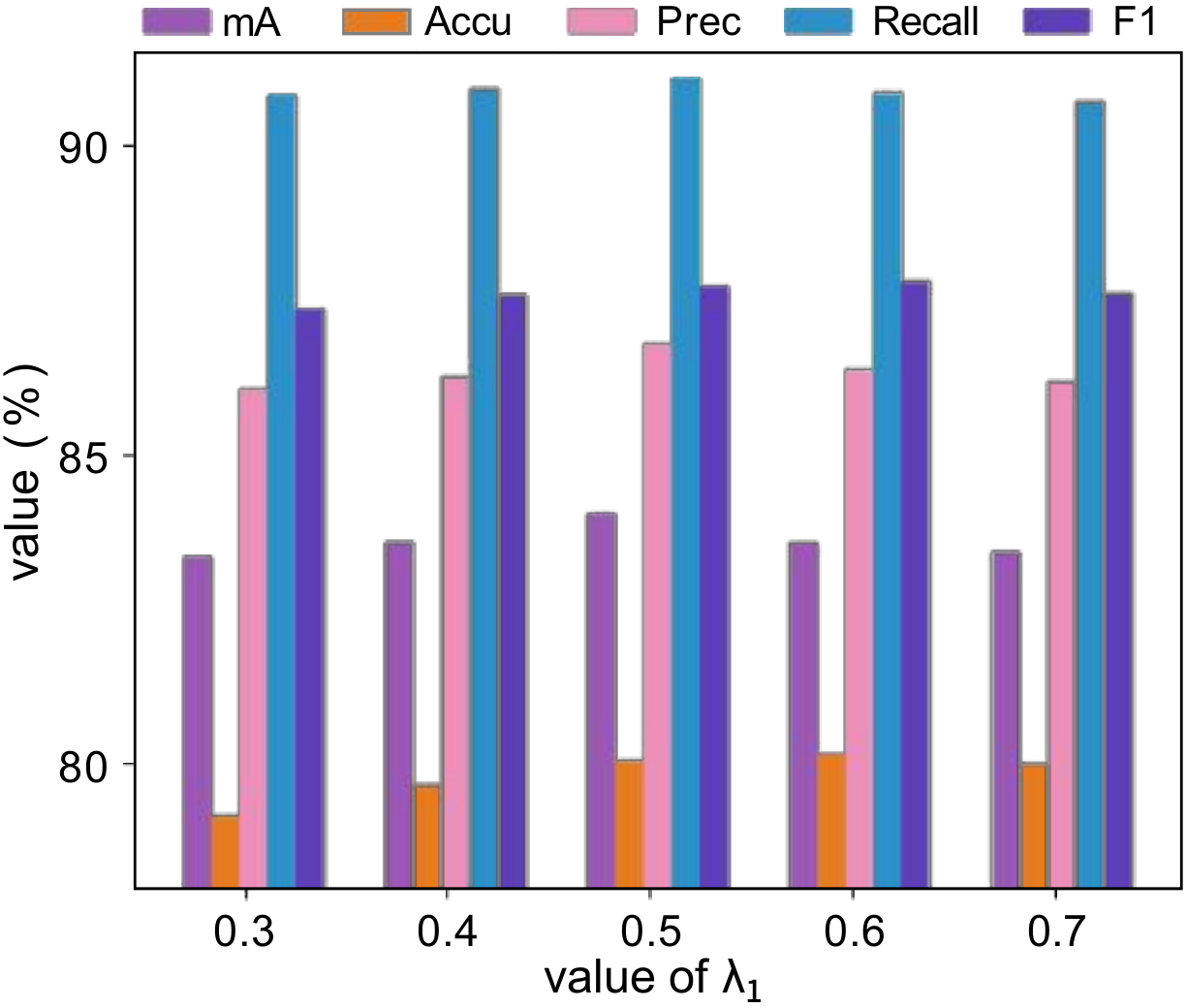}}
	\end{minipage} 
	\qquad 
	
	\begin{minipage}[t]{.3\linewidth}
		\centering
		\subfloat[][mask ratio on the PETA dataset]{\label{Genelecs:Genelec 8030 AP}\includegraphics[width=5.6cm]{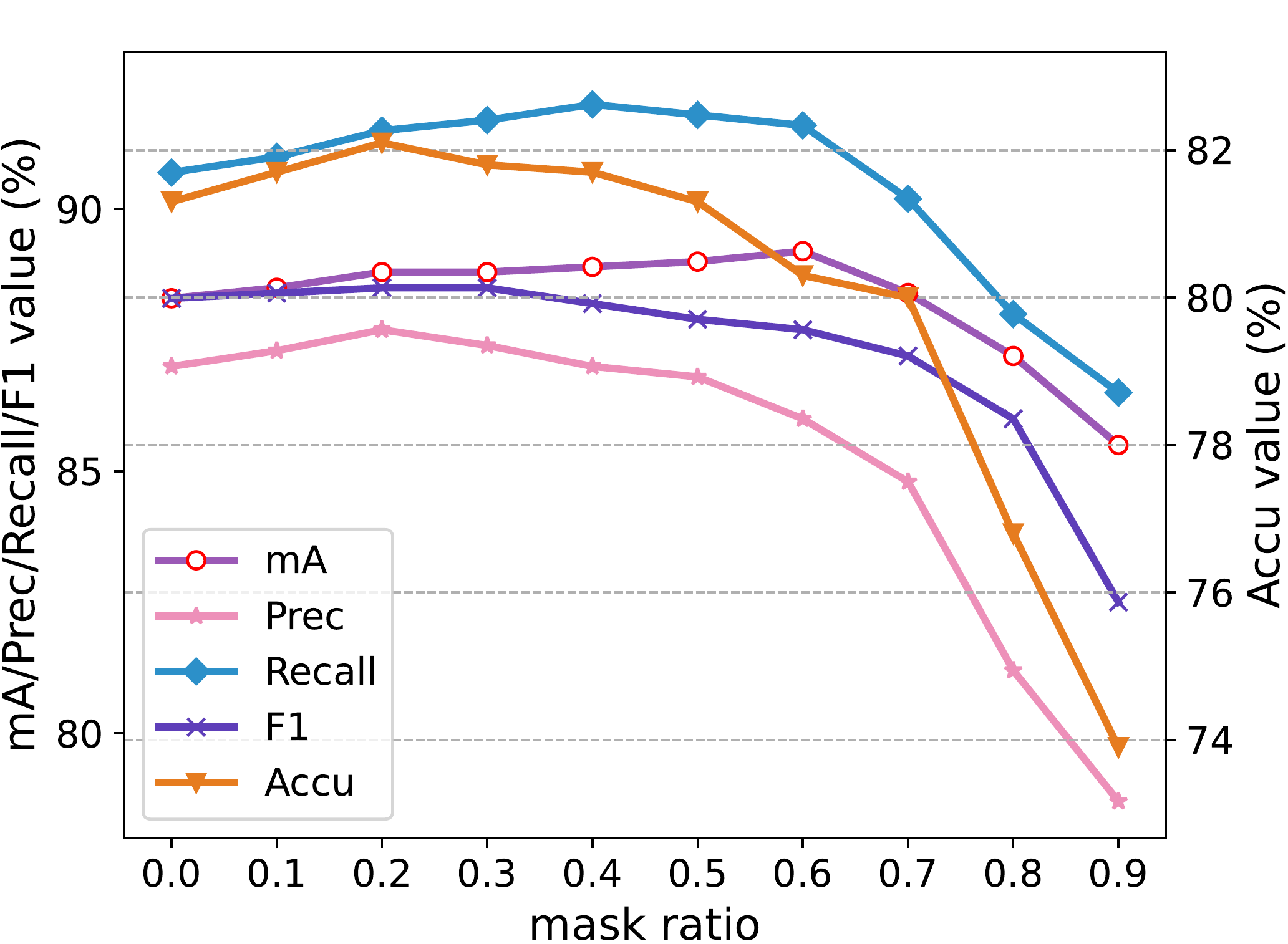}}
	\end{minipage} 
	\hspace{.15in}
	\begin{minipage}[t]{.3\linewidth}
		\centering
		\subfloat[][mask ratio on the RAP dataset]{\label{Genelecs:Genelec 8030 AP}\includegraphics[width=5.6cm]{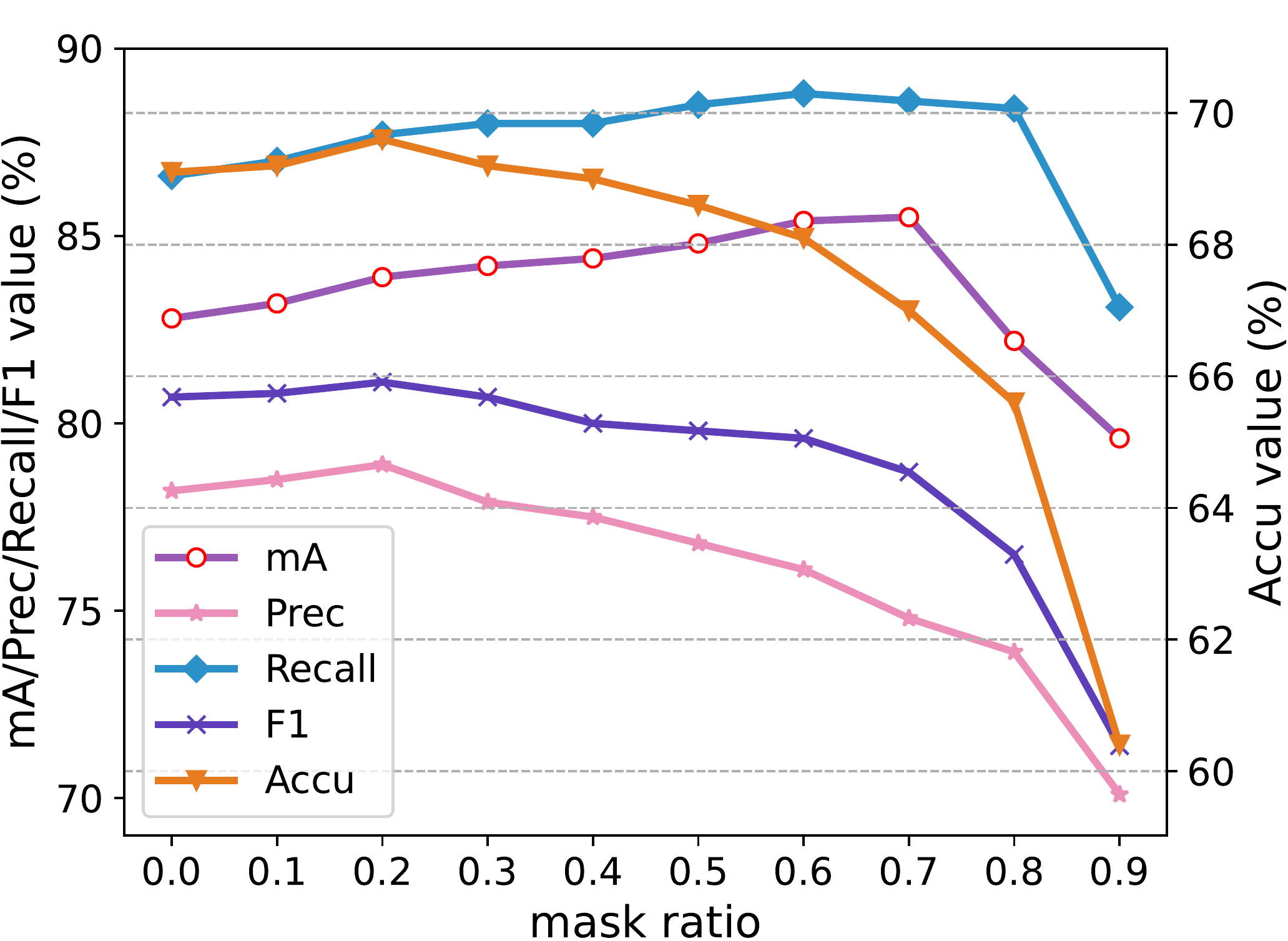}}
	\end{minipage} 
	\hspace{.15in}
	\begin{minipage}[t]{.3\linewidth}
		\centering
		\subfloat[][mask ratio on the PA100K dataset]{\label{Genelecs:Genelec 8030 AP}\includegraphics[width=5.6cm]{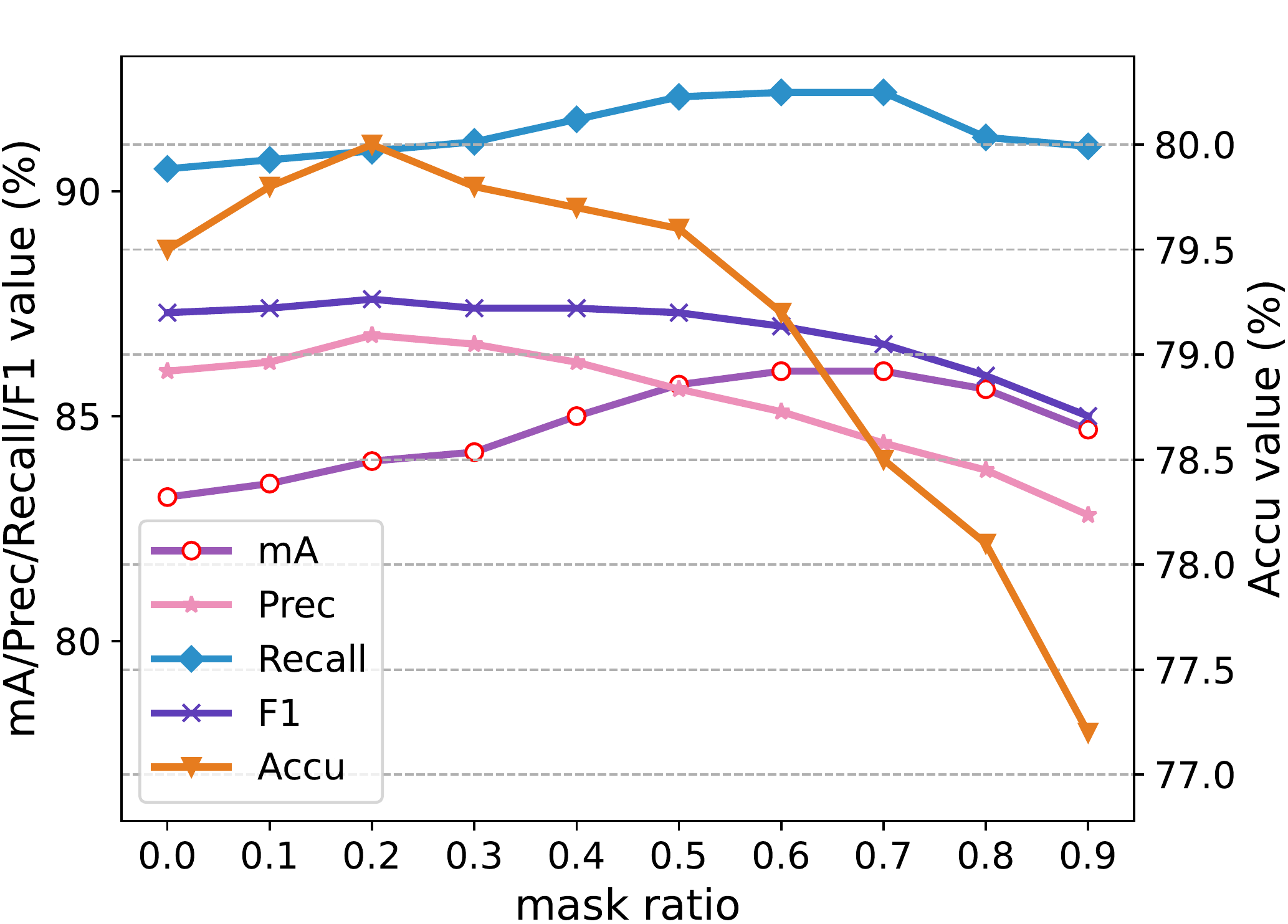}}
	\end{minipage} 
	\qquad  
	
	\begin{minipage}[t]{.3\linewidth}
		\centering
		\subfloat[][$T$ on the PETA dataset]{\label{Genelecs:Genelec 8030 AP}\includegraphics[width=5.6cm]{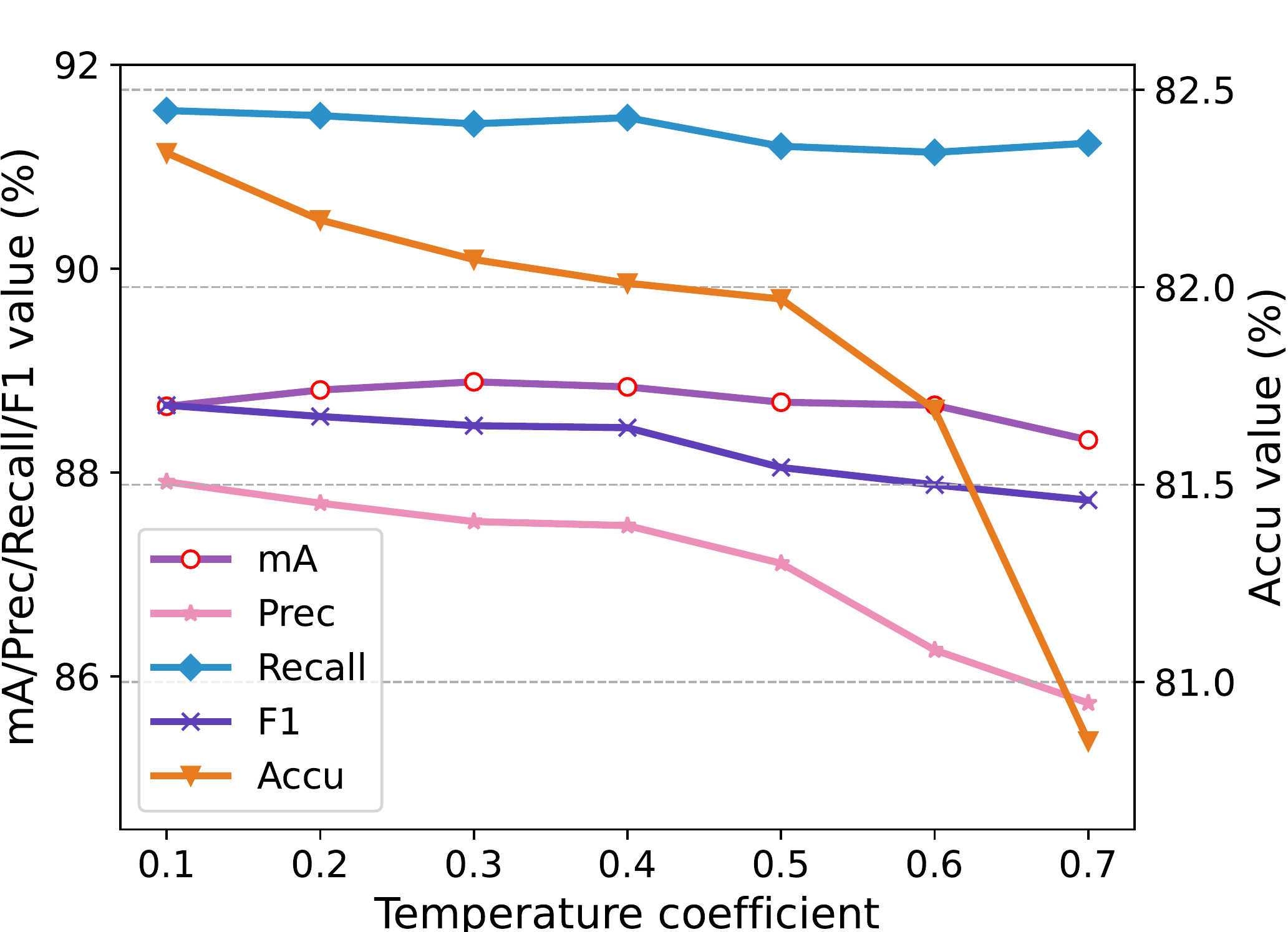}}
	\end{minipage} 
	\hspace{.15in}
	\begin{minipage}[t]{.3\linewidth}
		\centering
		\subfloat[][$T$ on the RAP dataset]{\label{Genelecs:Genelec 8030 AP}\includegraphics[width=5.6cm]{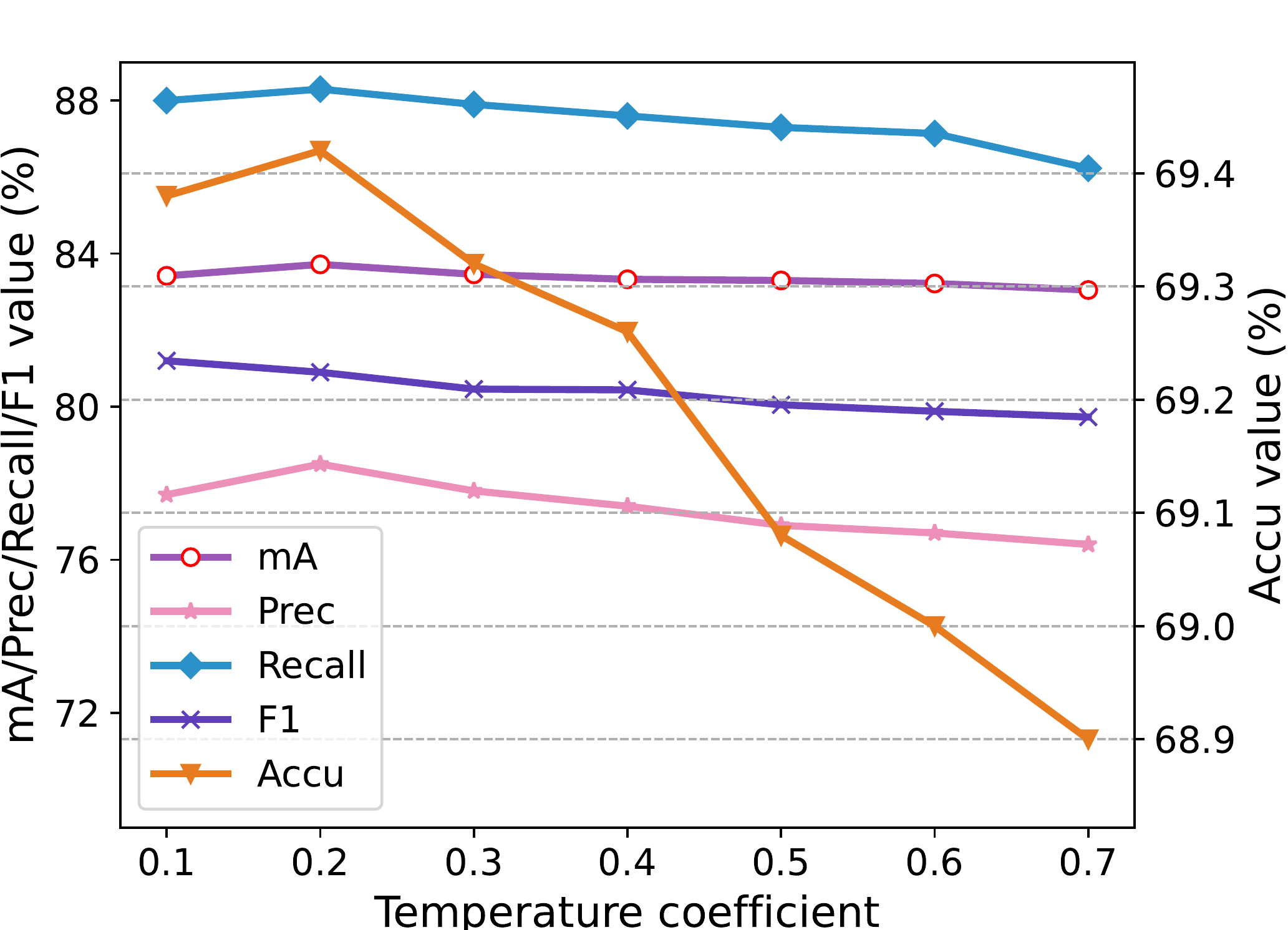}}
	\end{minipage} 
	\hspace{.15in}
	\begin{minipage}[t]{.3\linewidth}
		\centering
		\subfloat[][$T$ on the PA100K dataset]{\label{Genelecs:Genelec 8030 AP}\includegraphics[width=5.6cm]{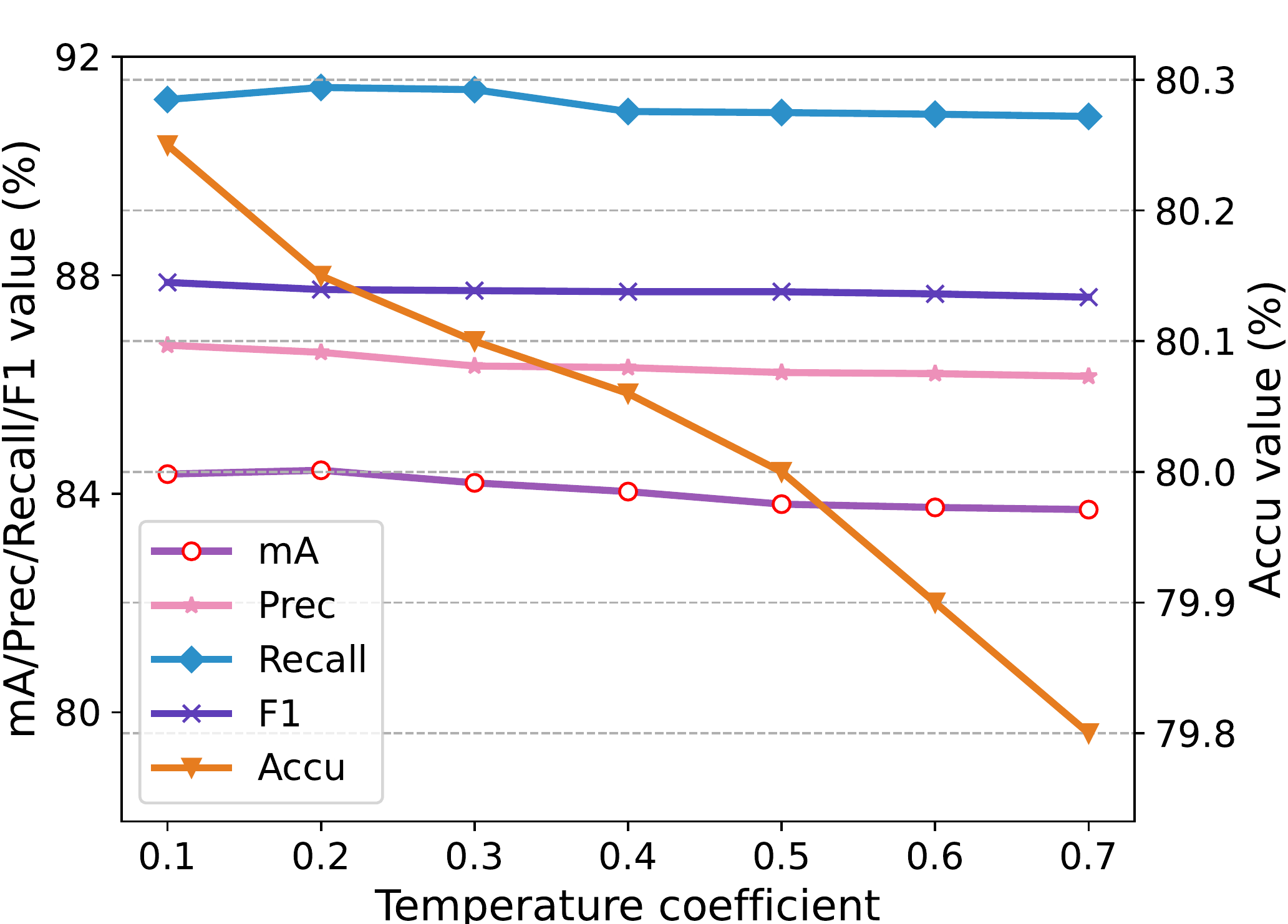}}
	\end{minipage} 
	\caption{Influence of $\lambda_1$, mask ratio, and $T$ on the performance of the proposed PARFormer network on the PETA, RAP, and PA100K datasets.}
	\label{fig:Genelecs}
\end{figure*}

\subsubsection{Analysis of the Mask Ratio}
The parameter mask ratio in the BRM block is used to control the masked probability of the feature maps. 
We conduct experiments to analysis the influence of the mask ratio on the performance of the BRM block. Fig. \textcolor{red}{7 (d)}, \textcolor{red}{(e)}, and \textcolor{red}{(f)} show how the mask ratio affects the performance on the three datasets. 
We observe that different metrics reach their peak at different mask ratios. With the increase of the mask ratio, the network performance is improved, but the metrics show a decreasing trend after reaching the peak. 
Excessive mask ratio will lead to a significant decline in the network performance. According to the performance results, we set the mask ratio to 0.3, 0.2, 0.2 on the PETA, RAP and PA100K datasets, respectively.

\subsubsection{Analysis of Multi-View Contrastive Loss}
In Equation (4), a temperature coefficient $T$ is used to control the attention applied to difficult negative samples. As $T$ decreases, more attention will be applied on the difficult negative samples, and greater gradient will be given to the difficult negative samples to separate them from the positive samples. 
To analyze the influence of MVCL on the recognition performance, we conduct ablation experiments under different values of $T$. Fig. \textcolor{red}{7 (g)}, \textcolor{red}{(h)}, and \textcolor{red}{(i)} show the experiment results on the three datasets. The performance of the PARFormer network is stable within a wide range for $T$ value. The network performs best when the $T$ is set to 0.1, 0.2, 0.1 on the PETA, RAP, and PA100K datasets, respectively. 

%% label ma
\begin{figure*}[t]
	\centering
	\includegraphics[height=4.8 cm]{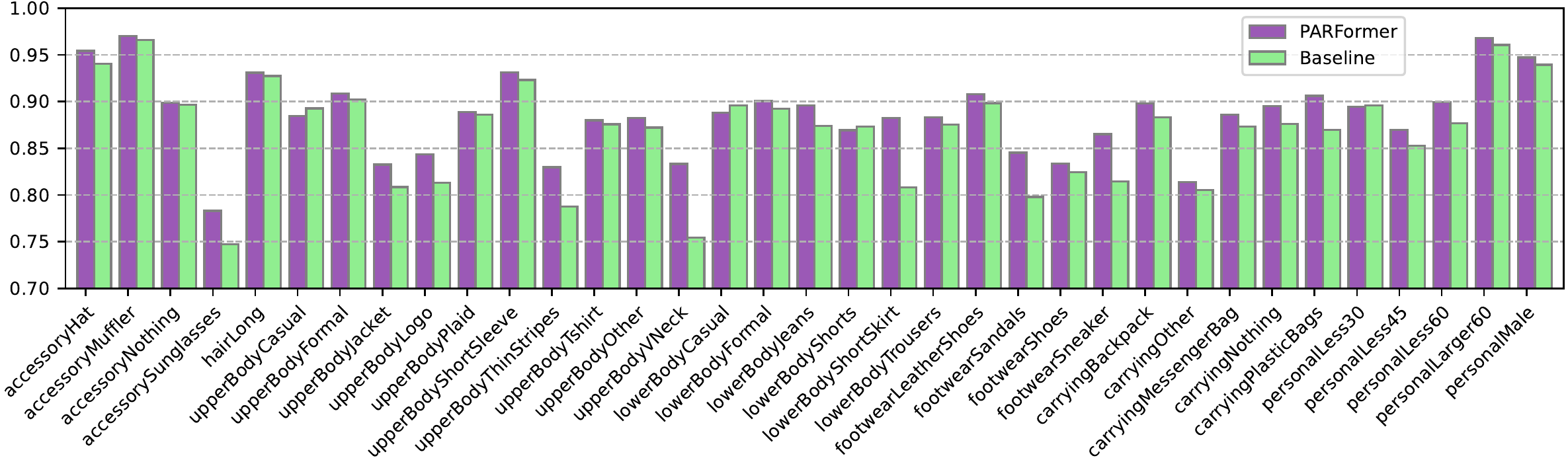}
	\caption{Attribute-wise mA comparison on the PETA dataset between the baseline network (green box) and the proposed PARFormer (purple box). PARFormer achieves significant improvements on most attributes.}
	\label{fig1}
	\centering
\end{figure*}

%% heat map
\begin{figure}[t]
	\centering
	\includegraphics[height=10.3 cm]{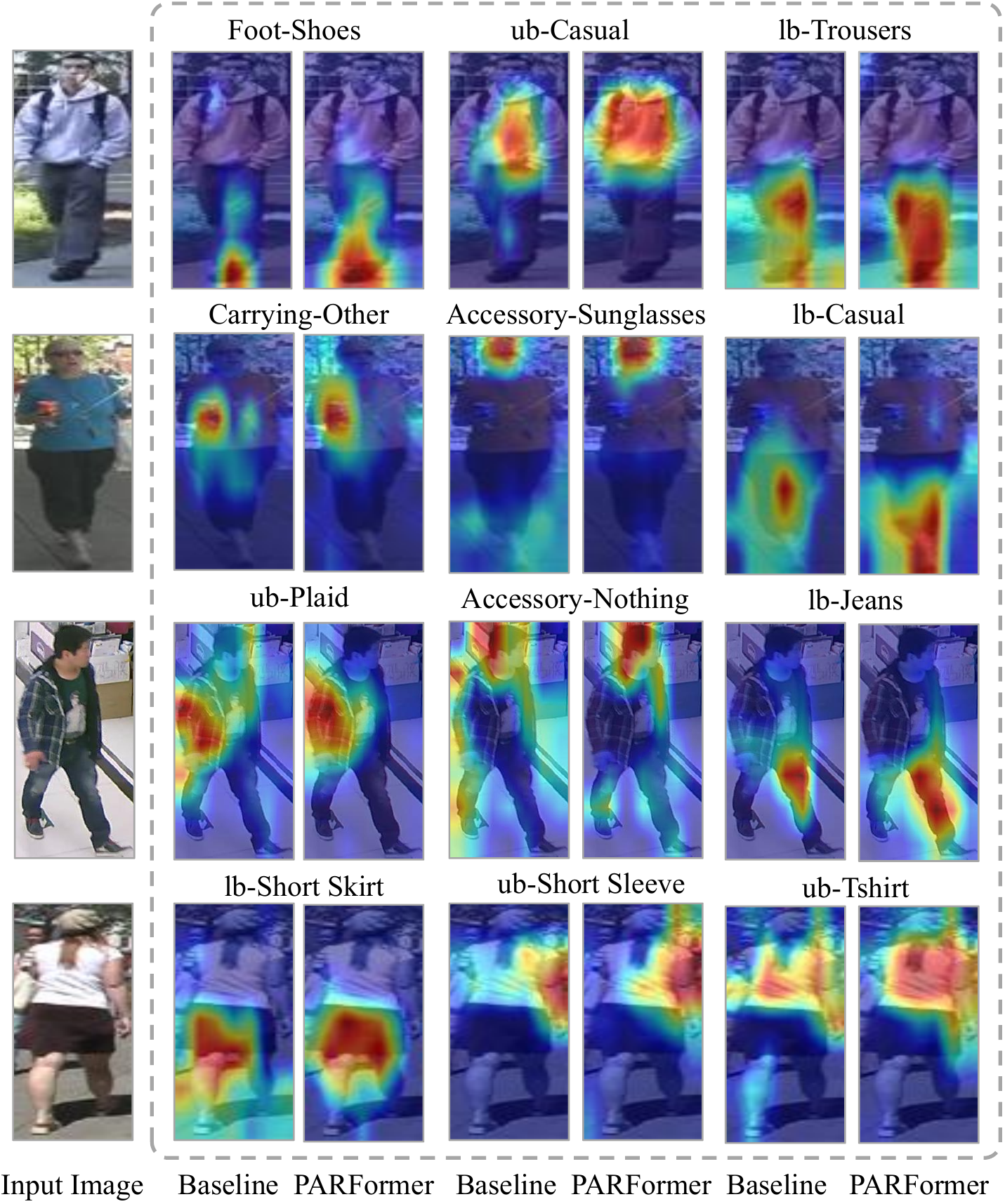}
	\caption{Grad-CAM visualization of the attention heat maps between the baseline and the proposed PARFormer, which confirms the effect of the proposed method. For the images in the first column, the other columns on the right side show the attention area of different attributes. }
	\label{fig1}
	\centering
\end{figure}

\subsubsection{Analysis of Final Results}
We compare the attribute-wise mA value between the baseline and the proposed PARFormer on the PETA dataset, which is shown in Fig. \textcolor{red}{8}. We can observe that PARFormer achieves significant improvement in many attributes, such as ‘accessorySunglasses', ‘upperBodyVNeck', ‘carryingBackpack’, etc. These attributes can be detected better at specific viewpoints. The results confirm that the viewpoint information can improve the accuracy of attribute recognition, and demonstrates the effectiveness of the proposed MVCL. 
The PARFormer also has obvious advantages in some fine-grained attributes, such as ‘upperBodyThinStripes’, ‘lowerBodyShortSkirt’, ‘footwearSneaker’, etc. These improvements may come from the effectiveness of the proposed feature processing module, which reinforces the attentive learning of the generated features. 

Fig. \textcolor{red}{9} shows the visualized specific-attribute attention heat maps for the comparison of the baseline and the proposed PARFormer network. 
The BRM block and MACL are effectively incorporated to further supervise the network training process, which makes the features extracted by the PARFormer network more discriminative and makes the network more robust to perturbations. 
It can be observed that the PARFormer network can discover more explicit part regions compared with the baseline network, and the regions neglected by the baseline network can be attentioned.

\section{Conclusion}
In this paper, we propose a novel transformer-based multi-task network PARFormer for the PAR task. 
We first introduce a BRM block to reinforce the attentive feature learning of the network and make it more adaptive to complex environments. 
To obtain more discriminative features, we propose a novel MACL to aggregate attributes to their respective centers and make them easier to be identified. 
Moreover, we propose a MVCL to enable the network to utilize the viewpoint information to assist the attribute recognition process, which facilitates the recognition of some viewpoint-specific attributes. 
With the incorporation of the proposed BRM block, MACL and MVCL into an end-to-end learning framework, the proposed PARFormer network can extract a highly discriminatory feature representations. 
The ablation studies show that each of these modules can bring performance improvements. 
Experimental results demonstrate that the proposed PARFormer network achieves superior performance against state-of-the-art PAR methods on multiple benchmark datasets. 
%This work shows the potential of the transformer-based method in the field of PAR. In the future, we can refer to the improvement strategy of the CNN method and propose a better transformer-based PAR framework.

%%%%%%%%%%%%%%%%%%%%%%%%%
%%%%%%%%%%%%%%%%%%%%%%%%%
%%%%%%%%%%%%%%%%%%%%%%%%%
%%%%%%%%%%%%%%%%%%%%%%%%%
%%%%%%%%%%%%%%%%%%%%%%%%%

\bibliographystyle{IEEEtran}
\bibliography{ref}

\vfill
\end{document}